\documentclass{article}

\PassOptionsToPackage{numbers, compress}{natbib}

\usepackage[main, final]{neurips_2026_vericode}

\makeatletter
\renewcommand{\@notice}{%
  \enlargethispage{2\baselineskip}%
  \@float{noticebox}[b]%
    \vspace*{-14pt}% <- adjust: more negative = smaller gap
    \footnotesize\@noticestring%
  \end@float%
}
\makeatother

\usepackage[utf8]{inputenc} % allow utf-8 input
\usepackage[T1]{fontenc}    % use 8-bit T1 fonts
\usepackage{hyperref}       % hyperlinks
\usepackage{url}            % simple URL typesetting
\usepackage{booktabs}       % professional-quality tables
\usepackage{amsfonts}       % blackboard math symbols
\usepackage{nicefrac}       % compact symbols for 1/2, etc.
\usepackage{microtype}      % microtypography
\usepackage{xcolor}         % colors
\usepackage{subcaption}
\usepackage{amsmath}
\usepackage{cleveref}
\usepackage{graphicx} 
\usepackage{wrapfig}

\newif\ifcomments\commentstrue   % \commentsfalse to hide all comments

\ifcomments
  \newcommand{\manna}[1]{\textcolor{purple}{\textbf{[Manna:} #1\textbf{]}}}
  \newcommand{\kseniia}[1]{\textcolor{orange}{\textbf{[Kseniia:} #1\textbf{]}}}
  \newcommand{\balazs}[1]{\textcolor{violet}{\textbf{[Balazs:} #1\textbf{]}}}
  \newcommand{\ferenc}[1]{\textcolor{green}{\textbf{[Ferenc:} #1\textbf{]}}}
  \newcommand{\note}[1]{\textcolor{orange}{\textbf{[note:} #1\textbf{]}}}
  \newcommand{\todoc}[1]{\textcolor{red}{\textbf{[TODO:} #1\textbf{]}}}
\else
  \newcommand{\manna}[1]{}
  \newcommand{\kseniia}[1]{}
  \newcommand{\balazs}[1]{}
  \newcommand{\ferenc}[1]{}
  \newcommand{\note}[1]{}
  \newcommand{\todoc}[1]{}
\fi

\newcommand{\draft}[1]{{\color{blue}\itshape #1}}
\usepackage{etoolbox}
\usepackage{enumitem}       % numbered lists with tight margins
\usepackage{float}          % [H] placement for the appendix figure
\usepackage{float}          % [H] placement for the appendix figure
\usepackage{xcolor}         % for \color in draftblock
\usepackage{amsmath}
\usepackage{amssymb}   % \triangleq, \rightsquigarrow, \Box, \Diamond in the TLA+ listings
\usepackage{tikz}
\usetikzlibrary{arrows.meta,positioning,calc,fit,backgrounds,shapes.geometric,shapes.misc}

\usepackage[most]{tcolorbox}
\usepackage{listings}

\definecolor{tlorange}{RGB}{230,118,34}   \definecolor{tlorangebg}{RGB}{253,238,221}
\definecolor{tlblue}  {RGB}{ 26,110,205}  \definecolor{tlbluebg}  {RGB}{220,234,250}
\definecolor{tlgreen} {RGB}{ 27,110, 88}  \definecolor{tlgreenbg} {RGB}{222,240,233}
\definecolor{tlred}   {RGB}{176, 42, 42}  \definecolor{tlredbg}   {RGB}{251,229,229}
\definecolor{tlsand}  {RGB}{193,138, 40}  \definecolor{tlsandbg}  {RGB}{253,246,227}
\definecolor{tlgrey}  {RGB}{125,125,125}  \definecolor{tlgreybg}  {RGB}{238,238,238}

\colorlet{tlgreenband}{tlgreenbg!62!white}
\colorlet{tlblueband} {tlbluebg!62!white}
\colorlet{tlredband}  {tlredbg!62!white}

\makeatletter
\@ifundefined{pgf@layerbox@tlback}{%
  \pgfdeclarelayer{tlback}%
  \pgfsetlayers{tlback,main}%
}{}
\makeatother

\tikzset{
  tlbox/.style={
    draw, rounded corners=3.5pt, line width=1.1pt,
    align=center, font=\small,
    inner xsep=7pt, inner ysep=6pt,
    minimum height=9mm,
  },
  tlorangebox/.style={tlbox, draw=tlorange, fill=tlorangebg},
  tlbluebox/.style  ={tlbox, draw=tlblue,   fill=tlbluebg},
  tlgreenbox/.style ={tlbox, draw=tlgreen,  fill=tlgreenbg},
  tlredbox/.style   ={tlbox, draw=tlred,    fill=tlredbg},
  tlsandbox/.style  ={tlbox, draw=tlsand,   fill=tlsandbg},
  tlgreybox/.style  ={tlbox, draw=tlgrey,   fill=tlgreybg},
  tlsquare/.style={tlbox, rounded corners=1pt},
  tlcircle/.style={
    draw, circle, line width=1.1pt, font=\small,
    minimum size=9mm, inner sep=1pt,
  },
  tlbluecircle/.style ={tlcircle, draw=tlblue,  fill=tlbluebg},
  tlredcircle/.style  ={tlcircle, draw=tlred,   fill=tlredbg},
  tlgreycircle/.style ={tlcircle, draw=tlgrey,  fill=tlgreybg},
  tlorangecircle/.style={tlcircle, draw=tlorange, fill=tlorangebg},
  tlfilebox/.style={
    align=center, font=\small, inner xsep=8pt, inner ysep=7pt,
    minimum height=9mm,
    append after command={\pgfextra{%
      \begin{pgfonlayer}{tlback}
        \path[draw=tlgrey, line width=1.1pt, fill=tlgreybg]
          (\tikzlastnode.south west) -- (\tikzlastnode.north west)
          -- ($(\tikzlastnode.north east)+(-2.8mm,0)$)
          -- ($(\tikzlastnode.north east)+(0,-2.8mm)$)
          -- (\tikzlastnode.south east) -- cycle;
        \path[draw=tlgrey, line width=1.1pt, fill=white]
          ($(\tikzlastnode.north east)+(-2.8mm,0)$)
          -- ($(\tikzlastnode.north east)+(-2.8mm,-2.8mm)$)
          -- ($(\tikzlastnode.north east)+(0,-2.8mm)$) -- cycle;
      \end{pgfonlayer}}},
  },
}

\tikzset{
  tlflow/.style={
    -{Triangle[width=5.5pt,length=5pt]}, line width=1.1pt, draw=black,
  },
  tlfeedback/.style={
    -{Triangle[open,width=6.5pt,length=6pt]}, line width=1.1pt,
    dash pattern=on 4pt off 2.5pt,
  },
  tlrefused/.style={tlfeedback, draw=tlred, -{Triangle[width=5.5pt,length=5pt]}},
  tledge/.style={font=\scriptsize, align=center, inner sep=2.5pt},
  tlnote/.style={font=\scriptsize\itshape, align=center, inner sep=2pt,
                 text=black!70},
  tltool/.style={tledge, font=\scriptsize\ttfamily},
  tlband/.style={
    rounded corners=5pt, line width=0.9pt, dash pattern=on 3pt off 2pt,
  },
  tlbandlabel/.style={font=\scriptsize\bfseries\ttfamily, inner sep=3pt},
  tlstep/.style={circle, draw=black, fill=white, line width=0.8pt,
                 inner sep=0pt, minimum size=4.4mm, font=\scriptsize\bfseries},
  tlxmark/.style={cross out, draw=tlred, line width=1.1pt,
                  minimum size=3.2mm, inner sep=0pt, fill=none},
}

\newcommand{\tlrole}[1]{{\footnotesize\bfseries\scshape #1}}

\tcbset{
  tlquote/.style={
    enhanced, breakable,
    colback=#1bg, colframe=#1, boxrule=1.1pt,
    arc=3.5pt, outer arc=3.5pt,
    left=7pt, right=7pt, top=6pt, bottom=6pt,
    fontupper=\small,
    coltitle=#1, colbacktitle=#1bg,
    titlerule=0pt, toptitle=1pt, bottomtitle=0pt,
    fonttitle=\scriptsize\bfseries\scshape,
  },
}

\newtcolorbox{proverquote}[1][]{tlquote=tlgreen,
  IfValueTF={#1}{title={#1}}{}}
\newtcolorbox{reviewerquote}[1][]{tlquote=tlred,
  IfValueTF={#1}{title={#1}}{}}
\newtcolorbox{gatequote}[1][]{tlquote=tlblue,
  IfValueTF={#1}{title={#1}}{}}

\lstdefinelanguage{TLA}{
  morekeywords={
    MODULE,EXTENDS,CONSTANT,CONSTANTS,VARIABLE,VARIABLES,ASSUME,THEOREM,
    LOCAL,INSTANCE,LET,IN,IF,THEN,ELSE,CASE,OTHER,CHOOSE,DOMAIN,SUBSET,UNION,
    EXCEPT,UNCHANGED,ENABLED,WF,SF,TRUE,FALSE,BOOLEAN,Nat,Int,Seq,Len,Append,
    Head,Tail,Cardinality},
  morekeywords=[2]{Init,Next,Spec,Send,Agreement,StartPreVote,AppendEntries},
  sensitive=true,
  morecomment=[l]{\\*},
  morecomment=[s]{(*}{*)},
  morestring=[b]",
  literate=
    {==}{{$\;\triangleq\;$}}3
    {/\\}{{$\land$}}1  {\\/}{{$\lor$}}1
    {~>}{{$\rightsquigarrow$}}2 {=>}{{$\Rightarrow$}}2
    {\\A}{{$\forall$}}1 {\\E}{{$\exists$}}1 {\\in}{{$\in$}}1
    {\\cup}{{$\cup$}}1 {\\cap}{{$\cap$}}1 {\\subseteq}{{$\subseteq$}}1
    {<>}{{$\Diamond$}}1 {[]}{{$\Box$}}1
    {|->}{{$\mapsto$}}2
    {∀}{{$\forall$}}1 {∃}{{$\exists$}}1 {∈}{{$\in$}}1
    {∧}{{$\land$}}1 {∨}{{$\lor$}}1 {⟹}{{$\Longrightarrow$}}2
    {□}{{$\Box$}}1 {◇}{{$\Diamond$}}1 {⟨}{{$\langle$}}1 {⟩}{{$\rangle$}}1,
}

\lstdefinelanguage{Verus}{
  morekeywords={
    as,break,const,continue,crate,else,enum,extern,false,fn,for,if,impl,in,
    let,loop,match,mod,move,mut,pub,ref,return,self,Self,static,struct,super,
    trait,true,type,unsafe,use,where,while,dyn},
  morekeywords=[3]{
    spec,proof,exec,requires,ensures,invariant,decreases,assert,assume,
    forall,exists,implies,open,closed,uninterp,broadcast,tracked,ghost,
    verus,by,via,recommends,opens_invariants,no_unwind},
  morekeywords=[2]{
    always,eventually,leads_to,weak_fairness,strong_fairness,enabled,
    lift_state,lift_action,tla_forall,valid,entails,wf1,sf1,leads_to_trans,
    StatePred,ActionPred,TempPred,Execution},
  sensitive=true,
  morecomment=[l]{//},
  morecomment=[s]{/*}{*/},
  morestring=[b]",
}

\lstdefinestyle{tlcode}{
  basicstyle=\ttfamily\scriptsize,
  keywordstyle=\color{tlblue}\bfseries,
  keywordstyle=[2]\color{tlgreen}\bfseries,
  keywordstyle=[3]\color{tlsand!75!black}\bfseries,
  commentstyle=\color{tlgrey}\itshape,
  stringstyle=\color{tlred!80!black},
  showstringspaces=false,
  columns=fullflexible,
  keepspaces=true,
  breaklines=true,
  breakatwhitespace=true,
  extendedchars=true,
  upquote=true,
  aboveskip=0pt, belowskip=0pt,
  numbers=none,
  escapeinside={(*@}{@*)},
}

\newtcblisting{tlacode}[1][]{
  tlquote=tlsand,
  listing only,
  listing options={style=tlcode, language=TLA},
  #1
}
\newtcblisting{veruscode}[1][]{
  tlquote=tlsand,
  listing only,
  listing options={style=tlcode, language=Verus},
  #1
}

\newtcbox{\tlinline}{on line, tlquote=tlsand, boxrule=0.8pt,
  left=2pt, right=2pt, top=1pt, bottom=1pt, arc=2pt,
  fontupper=\ttfamily\footnotesize}
\providecommand{\tlaplus}{TLA\textsuperscript{+}}
\providecommand{\code}[1]{\texttt{#1}}
\providecommand{\draftmark}{\textbf{[DRAFT]}~}
\providecommand{\draft}[1]{\begin{draftblock}#1\end{draftblock}}
\ifcsname draftblock\endcsname\else
  \newenvironment{draftblock}{\par\begingroup\color{blue}\itshape}{\endgroup\par}
\fi
\usepackage{pgfplots}
\pgfplotsset{compat=1.18}
\usepgfplotslibrary{groupplots}
\usepackage{xcolor}
\definecolor{llm}{HTML}{2A78D6}
\definecolor{alg}{HTML}{EB6834}
\definecolor{ink}{HTML}{0B0B0B}
\definecolor{gridc}{HTML}{E1E0D9}
\pgfplotsset{
  base/.style={grid=major, grid style={gridc, thin}, axis line style={gray!60},
    tick style={draw=none}, tick label style={font=\footnotesize}, label style={font=\small},
    title style={font=\small, align=left}, legend style={font=\footnotesize, draw=none, fill=none}, legend cell align=left},
  paired/.style={base, width=8.4cm, height=8.4cm, log ticks with fixed point},
  llmmark/.style={only marks, mark=*, mark size=1.3pt, llm, opacity=0.7},
  algmark/.style={only marks, mark=*, mark size=1.3pt, alg, opacity=0.7},
}

\title{TLA-Verus: Generating and Solving Temporal Proof Obligations with LLM Agents}

\author{%
  Kseniia Strelbytska\thanks{Work done while an intern at Reasonable. \quad $\dagger$ Equal contribution.}\hspace{0.4em}$^\dagger$\\
  Stephen Perse Cambridge, UK \\
  \texttt{kseniia@reasonable.io} \\
   \And
   Balázs Szilágyi$^{*\dagger}$ \\
   University of Cambridge, UK \\
   \texttt{balazs@reasonable.io} \\
   \AND
   Anna Mészáros$^*$ \\
   University of Cambridge, UK \\
   \texttt{manna@reasonable.io} \\
   \And
   Ferenc Huszár \\
   Reasonable \\
   \texttt{ferenc@reasonable.io} \\
}

\begin{document}

\maketitle

\begin{abstract}
  The TLA+ (temporal logic of actions) formal specification toolkit has seen widespread adoption in modeling modern distributed software systems including databases, storage systems, ledgers and coordination protocols. This popularity is despite its two significant limitations for software verification: Firstly, TLA+ offers no way to detect drift between software implementation and mathematical specification; secondly, TLC and TLAPS cannot provide satisfactory proofs of arbitrary temporal logic properties. We address the second limitation and take a first step towards the first by translating TLA+ specifications into Verus, an auto-active verifier built for Rust, via both a deterministic transpiler and an agentic translator. We demonstrate that our LLM-based pipeline is able to generate proof obligations from TLA+ specifications at scale, and to produce fully machine checkable proofs of both safety and liveness properties. Since temporal logic reasoning in Verus is a largely new task for LLMs, our work yields a valuable evaluation dataset, which we use to map out current generation model capabilities in Verus theorem proving.
  
\end{abstract}

\section{Introduction}
\label{sec:intro}
\vspace{-0.5em}
Frontier coding agents show remarkable success in automating complex proof search, solving one of the most challenging technical problems in software verification. However, mainstream adoption of formal verification in modern software is not exclusively a technical capability question: it depends on how well verification integrates into the workflows developers already use, and on finding repeatable ways of specifying system properties — general enough languages in which idiosyncratic correctness properties of many different systems can be expressed compactly.

TLA+ is a successful example of a formal modeling toolkit that ticks these boxes. It is widely deployed in modern, mainstream distributed systems engineering, see e.\,g.\ MongoDB \cite{davis2020extreme}, Amazon Web Services \cite{newcombe2015aws,newcombe2014why}, Kafka's KRaft protocol \cite{vanlightly2024kraft}, or Datadog \cite{parthiban2024courier}, etc. In the modern context TLA+ is dated technology, and suffers from two main weaknesses:
\begin{itemize}
\vspace{-0.5em}
    \item there is no way to prove that the implementation refines the mathematical specification, often leading to implementation - specification drift,
    \vspace{-0.5em}
    \item limited support for proofs especially of liveness properties for large, complex systems.
\end{itemize}
\vspace{-0.5em}

Temporal logic reasoning has more recently been embedded directly into auto-active deductive verifiers, allowing liveness proofs to live alongside safety proofs in the same verified codebase rather than in a separate model. Microsoft's IronFleet project pioneered this for a Paxos-based replication protocol\ \cite{hawblitzel2015ironfleet}, and more recently, Anvil embedded temporal logic reasoning in Verus to verify liveness of Kubernetes cluster management controllers \cite{sun2024anvil, cai2026welder}.

This work builds directly on Anvil, and significantly extends the number of temporal logic proofs carried out in this style. It addresses the second weakness and takes a step towards the first: we bring specifications into Verus, where an implementation can later be proved to match them.

Beyond the value this provides to users of temporal logic directly, it also provides value for the development of better vericoding agents: it yields a new dataset of safety and liveness proofs that is not academic in size or focus, but practically motivated — and, critically, one whose solutions do not already appear in the training data of existing systems.

The contributions of this work are:
\vspace{-1em}
    \begin{itemize}
        \item we describe and compare two approaches for translating TLA+ specifications to Verus: a purely algorithmic transpiler, and one based on an agentic workflow (\Cref{sec:phase1})
        \item we describe a multi-turn, prover-reviewer pipeline for writing proofs of TLA properties (\Cref{sec:pahse2})
        %\item we successfully proved XXXX properties using Anthropic's Opus, and YYYY properties using the open-weight GLM-5.3 \manna{3094 tasks were proven for the algorithmic translations, and 492 to the agentic one, but I don't know which used which model.}
        \vspace{-0.5em}
        \item we successfully prove 3,094 tasks' algorithmic translations, and 492 tasks' agentic translations using Anthropic's Opus. Moreover, we select 40 tasks and build two datasets from them: \texttt{tlaverus\_algo} from their algorithmic translations and \texttt{tlaverus\_llm} from their agentic ones (\Cref{sec:dataset})
        \vspace{-0.5em}
        \item we compare proof search across the two encodings by running the prover--reviewer pipeline on the same 500 (specification, property) pairs, translated both algorithmically and agentically (\Cref{sec:agentvsalg})
        \vspace{-0.5em}
        \item we evaluate a range of state-of-the-art language models on three datasets: \texttt{tlaverus\_algo}, \texttt{tlaverus\_llm} and \texttt{verusbench} (\citep{autoverus}) (\Cref{sec:modeleval}).
    \end{itemize}

%% GENERATED from docs/paper-section-prover-reviewer.tex by split-section.py -- edit THAT file.
\begin{figure}[H]
  \centering
  \resizebox{\textwidth}{!}{% Figure 1: the pipeline.  Translation is one of two arms; the Verus bundle is
% a dog-eared file holding its four parts, stacked in one column top to
% bottom: TLA+ library, model spec, property, theorem stub; proof search is a
% rectangular cycle (prover -> cheat checks -> reviewer -> gate) in one blue,
% light at the prover and darkest at the decision.  "no" closes the cycle;
% "yes" leaves it to the right as a merged proof or a finding.
%
% Every arrow is orthogonal and every endpoint is a node anchor.  Every box is
% two lines, so the boxes are the same height.  Words that are not on the
% critical path (which translator, what the checks are) belong in the
% caption.  Wider spacing between stages than the v1 layout; the bundle is a
% single narrow column rather than a 2x2 grid, and labels are trimmed.
\begin{tikzpicture}[
  tlsub/.style={tlgreybox, draw=tlgrey!45!white, line width=0.8pt,
                font=\scriptsize, text width=20mm,
                inner xsep=3pt, inner ysep=2.5pt, minimum height=6mm},
  tldiamond/.style={diamond, draw=tlblue!75!black, fill=tlblue!48!white, line width=1.1pt,
                    aspect=2.4, inner sep=1pt, font=\small, align=center},
  tlortho/.style={rounded corners=2pt},
  tlheading/.style={font=\footnotesize\bfseries\scshape, inner sep=2pt},
  % the loop, light to dark
  tlb1/.style={tlbox, draw=tlblue!45!white, fill=tlbluebg!40!white},
  tlb2/.style={tlbox, draw=tlblue!70!white, fill=tlbluebg!80!white},
  tlb3/.style={tlbox, draw=tlblue!90!black, fill=tlblue!25!white},
]

  % ---- input --------------------------------------------------------------
  \node[tlfilebox, text width=16mm] (S) at (0,0)
        {TLA\textsuperscript{+}\\spec};

  % ---- translation: two arms under one label ------------------------------
  \node[tlorangebox, text width=18mm] (TA) at (3.6, 0.75) {algorithmic};
  \node[tlorangebox, text width=18mm] (TB) at (3.6,-0.75) {agentic};
  \node[tlheading, text=tlorange, anchor=south] at (TA.north) {translation};
  \node[font=\small\itshape] at ($(TA.south)!0.5!(TB.north)$) {or};

  % ---- the Verus bundle: a dog-eared file, one column, top to bottom ------
  \node[tlsub] (B1) at (8.0, 1.2)  {TLA\textsuperscript{+}\\library};
  \node[tlsub] (B2) at (8.0, 0.4)  {model\\spec};
  \node[tlsub] (B3) at (8.0,-0.4)  {property};
  \node[tlsub] (B4) at (8.0,-1.2)  {theorem\\stub};
  \node[inner sep=1pt] (BT) at (8.0, 1.85) {\tlrole{verus bundle}};
  \node[inner xsep=5pt, inner ysep=4pt, fit=(B1)(B2)(B3)(B4)(BT)] (B) {};
  \begin{pgfonlayer}{tlback}
    \path[draw=tlgrey, line width=1.1pt, fill=white]
      (B.south west) -- (B.north west)
      -- ($(B.north east)+(-3.2mm,0)$) -- ($(B.north east)+(0,-3.2mm)$)
      -- (B.south east) -- cycle;
    \path[draw=tlgrey, line width=1.1pt, fill=tlgreybg]
      ($(B.north east)+(-3.2mm,0)$) -- ($(B.north east)+(-3.2mm,-3.2mm)$)
      -- ($(B.north east)+(0,-3.2mm)$) -- cycle;
  \end{pgfonlayer}

  % ---- proof search loop: prover on the axis, the cycle rising above ------
  \node[tlb1, text width=27mm] (P) at (13.5,-0.875)
        {\tlrole{prover}\\writes proof};
  \node[tlb2, text width=27mm] (C) at (13.5, 0.875)
        {\tlrole{cheat checks}\\no reward hacking};
  \node[tlb3, text width=27mm] (R) at (18.8, 0.875)
        {\tlrole{reviewer}\\steering};
  \node[tldiamond] (G) at (18.8,-0.875) {both\\accept?};
  \node[tlheading, text=tlblue] (H) at ($(C.north)!0.5!(R.north)+(0,0.3)$)
        {proof search loop};

  % ---- outputs ------------------------------------------------------------
  \node[tlgreenbox, text width=27mm] (V) at (24.1, -0.175)
        {\tlrole{merged proof}\\machine-checked};
  \node[tlredbox,   text width=27mm] (D) at (24.1, -1.575)
        {\tlrole{finding}\\property false};

  % ---- flow, all orthogonal -----------------------------------------------
  \draw[tlflow, tlortho] (S.east)  -- ++(0.4,0) |- (TA.west);
  \draw[tlflow, tlortho] (S.east)  -- ++(0.4,0) |- (TB.west);
  \draw[tlflow, tlortho] (TA.east) -- ++(0.4,0) |- (B.west);
  \draw[tlflow, tlortho] (TB.east) -- ++(0.4,0) |- (B.west);
  \draw[tlflow, tlortho] (B.east) -- ++(0.45,0) |- (P.west);

  % the cycle, clockwise from the prover
  \draw[tlflow] (P.north) -- (C.south);
  \draw[tlflow] (C.east)  -- (R.west);
  \draw[tlflow] (R.south) -- (G.north);
  \draw[tlfeedback, draw=tlblue] (G.west) -- (P.east)
        node[tledge, pos=0.5, below=3pt] (N) {\textbf{no:}\\next round};

  % yes: out to the right
  \draw[tlflow, tlortho, draw=tlgreen] (G.east) -- ++(0.5,0) |- (V.west);
  \draw[tlflow, tlortho, draw=tlred]   (G.east) -- ++(0.5,0) |- (D.west);
  \node[tledge, anchor=south] at ($(G.east)+(0.28,0.1)$) {\textbf{yes}};

  % a pale outline that sets the loop apart without boxing it in
  \begin{pgfonlayer}{tlback}
    \node[tlband, draw=tlblue!40!white, fill=tlbluebg!35!white,
          inner xsep=7pt, inner ysep=6pt, fit=(P)(C)(R)(G)(H)(N)] {};
  \end{pgfonlayer}
\end{tikzpicture}}
  \caption{The pipeline end to end. A \tlaplus{} specification becomes a Verus
  bundle with one property to prove; the merge gate turns the bundle into either
  a merged, machine-checked proof or a finding about the specification itself.}
  \label{fig:pipeline}
\end{figure}
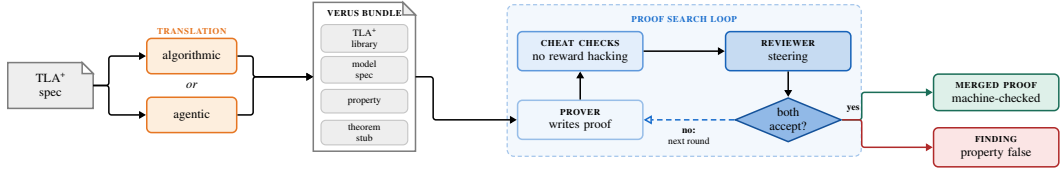

\section{Background}
\vspace{-0.5em}
\subsection{Temporal logic and TLA+ }
\label{sec:tla}
\vspace{-0.5em}
%% GENERATED from docs/paper-section-background.tex by split-background.py -- edit THAT file.
Software systems change their state in response to various events. Claims about the system's safe or correct behavior can be formalized as predicates over execution traces, sequences of states hypothetically visited by the system. Temporal logic of actions~\citep{pnueli1977temporal}, a variant of linear temporal logic, allows us to reason about such temporal predicates, and provides a language to describe behavioural properties of interest. In TLA+ we use three operators to construct temporal properties: $\Box P$ (always, $P$ holds in every state visited), $\Diamond P$ (eventually, $P$ holds in some state) and
$P \leadsto Q \equiv \Box(P \Rightarrow \Diamond Q)$ ($P$ leads to $Q$, whenever $P$ holds, eventually $Q$ will hold at some point in the future). In this language one can express a range of properties, the most typical ones can be categorised as:
\vspace{-0.5em}
\begin{description}
    \item[safety ($\Box \neg \text{Something Bad}$):] undesired states (incorrect, inconsistent, deadlocked, etc) are avoided.
    \item[liveness ($\Diamond \text{Something Good}$):] a desired state is eventually reached.
\end{description}
\vspace{-0.5em}
TLA+~\citep{lamport1994tla,lamport2002specifying} describes a system as a temporal
formula that holds for all executions that the system can produce:
\[
  \mathit{Spec} \;\equiv\; \mathit{Init} \wedge \Box[\mathit{Next}]_{\mathit{vars}} \wedge F\]
  
$\mathit{Init}$ declares what is true at initialisation. $\mathit{Next}$ is a binary relation between a previous and next state, which defines the state transitions possible. $\mathit{Next}$ itself is defined as a union of named relations called actions. Grouping transitions to named actions serve easier interpretation and play a role in defining fairness, as explained below. The $\Box[\mathit{Next}]_{\mathit{vars}}$ notation means that
every step either satisfies $\mathit{Next}$ or keeps all variables $\mathit{vars}$
unchanged, called a stutter step. Stutter steps allow the composition of multiple TLA+ modules: steps where variables don’t change allow variables in other components of a large system to change asynchronously.

These two conjuncts ($\mathit{Init}$ and $\mathit{Next}$) define what the system \emph{can} do, but not what it is \emph{likely} to do or what it \emph{must} do to make progress. In this sense they are too weak to imply correct behaviour. To reason about liveness properties, we want to rule out "degenerate" execution traces where nothing happens, or where only irrelevant and uninteresting events keep happening in a loop. This is why we include fairness assumptions $F$. These are individually defined over actions and require that if an action $A$ stays enabled, then it is
eventually taken. Specs differ in whether they assume  weak $\mathrm{WF}(A)$ or strong ($\mathrm{SF}(A)$) on actions that make progress.

TLA+ is widely used in industry to this day \cite[see e.\,g.\,][]{lamport2025industrial, newcombe2015amazon, davis2020extreme, parthiban2024courier} but has two major drawbacks:
\vspace{-0.5em}
\begin{description}
    \item[specification-implementation drift:] The specification and the implementation are separate artifacts in different languages, and the correspondence between them itself is not checkable. If the code changes, there is no obligation to revise the model, so a specification's guarantees might not hold for the code that implements it. In this sense, TLA+ is used only for model checking, rather than software verification.
    \item[limited proof support:] Properties declared in TLA+ can be checked either by TLC \citep{yu1999tlc}, an explicit state model checker, or TLAPS \citep{cousineau2012tlaps} an interactive theorem prover. Both have limitations: TLC cannot handle state space blow-up, thus its use is restricted to small configurations of systems, potentially missing errors that occur as systems are scaled up. TLAPS avoids explicit state-space exploration, but its support for liveness proofs is limited.
\end{description}

\vspace{-1em}
\subsection{Verus}
\label{sec:verus}
%% GENERATED from docs/paper-section-background.tex by split-background.py -- edit THAT file.

\begin{table}[b]
\centering
\small
\begin{tabular}{@{}ll@{}}
\toprule
\textbf{TLA} & \textbf{Verus (Anvil encoding)} \\
\midrule
execution trace                            & $\verb|Execution<State>| = \mathbb{N} \to \verb|State|$ \\
state predicate                            & $\verb|State| \to \verb|Bool|$\\
action relation                            & $\verb|State| \times \verb|State| \to \verb|Bool|$\\
temporal predicate                         & $\verb|Execution<State>| \to \verb|Bool|$\\
\midrule
state predicate $P$ holds at state 1       & \verb|lift_state(p)| \\
action predicate $A$ holds at first transition & \verb|lift_action(a)| \\
$\mathrm{WF}_{\mathit{vars}}(A)$, $\mathrm{ENABLED}\ A$
                                          & \verb|weak_fairness(a)|, \verb|enabled(a)| \\
$\mathit{Init} \wedge \Box[\mathit{Next}]$
                                          & \verb|and(lift_state(init), always(lift_action(next)))| \\
\midrule
$\Box P$, $\Diamond P$                    & \verb|always(p)|, \verb|eventually(p)| \\
$P \leadsto Q$                            & \verb|p.leads_to(q)| \\
$\mathit{Spec} \Rightarrow \Phi$          & \verb|spec().entails(phi)| \\
\bottomrule
\end{tabular}
\vspace{1em}
\caption{TLA+ concepts and operators used in specifying models and properties (\emph{left}) and their Verus encoding (\emph{right}), adapted from Anvil~\citep{sun2024anvil}. An execution trace is modeled as a Verus spec function mapping natural numbers to a state while predicates are encoded as functions returning a Boolean value. The operators `always` and `eventually` are implemented as Verus spec functions with `forall` and `exists` constructs, with control over SMT trigger selection. The temporal library also includes proven lemmas, such as the transitivity of $\leadsto$ and Lamport's WF1 rule \cite{lamport1994tla} for reasoning under a weak fairness assumption.}
\label{tab:verus-tla}
\end{table}

Verus is an auto-active verifier that unites specification, implementation, and proof in one language built on Rust \cite{lattuada2023verus}. While primarily designed to support contracts for methods, its rich ghost code support allows reasoning about abstract models of behaviour, so long as the mathematical content remains within the capabilities of an SMT solver.

In our context, Verus thus promises a fully end-to-end environment for verified code synthesis, addressing the specification-implementation gap in temporal logic: the TLA specification can be tied directly to the actual implementation by a refinement proof, as demonstrated for Kubernetes controllers in the Anvil library~\cite{sun2024anvil}. Furthermore, Verus supports fully machine checkable proofs for both safety and liveness properties.

In this paper, we focus on translating system specifications written in TLA+ to Verus (Section~\ref{sec:phase1}), and then using an agentic workflow to produce Verus-checked proofs of system properties (Section \ref{sec:pahse2}). Connecting implementations to the specifications via refinement proofs is left for future work.

% Developing systems in Verus introduces another problem that we address in this paper.

% \textbf{P2: proof cost.}
% Each line of Rust implementation requires, on average, 5 lines of proof code in Verus systems that ~\citep{lattuada2024verus} investigated.
% Writing proof annotations is slow, expert work,
% which holds formal verification back from mainstream software development.
% Proof generation is an attractive target for LLM assistance. Verus mechanically
% checks every proof it accepts, so a generated proof that verifies is as trustworthy as
% a human-written one, provided there are no cheating patterns (for instance, modifying the specification or adding \texttt{assume(false)}).

% In Sections~\ref{sec:pahse2} and~\ref{sec:modeleval}, we investigate P2 in two ways. Firstly, we propose a prover-reviewer system
%  for automatic proof generation in Verus. Secondly, we evaluate how well modern LLMs 
% perform on proof generation for Verus systems.

\section{The pipeline}
\label{sec:pipeline}
\vspace{-0.5em}
\subsection{Translation}
\label{sec:phase1}
\vspace{-0.5em}
The first step in our pipeline is to embed a TLA+ spec of an arbitrary system into Verus. For this embedding we use a temporal logic library based on that of Anvil \cite{sun2024anvil}. Table~\ref{tab:verus-tla} summarises the main code constructs used in this embedding.

This translation step is non-trivial because of the semantic gap between TLA+ and Verus. We developed two approaches to this problem which make different trade-offs: an algorithmic transpiler written in Rust we call \texttt{TranSpec}; and an agentic translation loop that uses an LLM agent paired with an LLM judge.
\vspace{-0.5em}
\subsubsection{Algorithmic Transpilation (\texttt{TranSpec})}
\vspace{-0.5em}
In this section we provide an overview of the key challenges our transpiler program  \texttt{TranSpec} has to overcome to translate TLA+ to valid Verus specifications.
\vspace{-1em}
\paragraph{Type System:}
The transpiler has to bridge the gap between untyped TLA+ and strongly typed Verus and Rust. \texttt{TranSpec} performs constraint solving, synthesizes enums, records, maps and sequences. The transpiler uses \texttt{Optional} types to handle TLA+'s untyped \texttt{CONSTANT NULL}, and `Sum` types to encode TLA+ variables that hold different types depending on system state.
\vspace{-1em}
\paragraph{Verus specific: termination measures, SMT triggers:} The transpiler has to produce valid Verus code, and that is a high bar. Recursive function calls in Verus require a termination measure (via \texttt{decreases} annotations), which have to be inferred by \texttt{TranSpec}. Every quantifier expression in the program (\texttt{forall}, \texttt{exists} or \texttt{choose}) should also include trigger annotations \cite{lattuada2023verus} which control how the SMT solver instantiates quantified variables. These triggers do not impact the semantics but determine the search space the solver explores: incorrect trigger annotations may lead to infinite loops or out-of-memory errors. Determining the correct placement of trigger annotations remains the most challenging task of this transpiler.
\vspace{-1em}
\paragraph{Ambiguous and unusual syntax:} TLA+ writes \texttt{f[x]} for functions, sequences and maps, and it is not always clear from the context where this appears which of these the creator of the spec intended. \texttt{TranSpec} delays this decision until further information is available. TLA+ includes other elements of unusual syntax that don't map cleanly to modern programming language syntax.
\vspace{-0.5em}
\subsubsection{Agentic Translation}
\vspace{-0.5em}
Even after solving the aforementioned challenges of mechanistic transpilation, an important problem remains: Some specs may contain false properties, or, more commonly, properties whose truth value is contingent on implicit assumptions not captured in the TLA+ spec. Implicit assumptions may lurk in a TLC config file or the way the model checker interprets specs and explores the state space.

To allow greater flexibility in \emph{creatively} fixing these issues, and to handle the long tail of TLA+ syntax, we also leverage LLM agents to perform the translation. The faithfulness of the translation is then checked by a reviewer LLM, resulting in a translator-reviewer loop similar to the prover-reviewer loop explained in Section\ \ref{sec:pahse2}. We ran experiments with both translation approaches, see quantitative comparison in \Cref{sec:agentvsalg}.

% Hand-written.  The four-category property taxonomy, moved here from Section 2.1
% on 2026-09-08: the labels are the dataset's, and the evaluation reports results
% by them, so they are defined once at the point where the counts appear.
% Wording is verbatim from paper-section-background.tex; only the four clauses
% were turned into list items.

% \begin{draftblock}
%     For agentic translation, we use a simplified version of the two-agent setup from ~\ref{sec:pahse2}. We remove the merge gate and anti-cheat checks, and require Verus to verify the bundle with assume() stubs instead of proofs. The translator agent writes the translation, and the reviewer agent catches errors. We use Claude Sonnet 5 for both agents. The translator is instructed to translate faithfully, is shown an example of the desired bundle structure (EWD840), and is shown mistakes from previous translations to avoid. The reviewer agent validates the translation against TLA+ and rejects a weakened specification, including if added assumptions are stronger than the TLA+. 
% \end{draftblock}
\vspace{-0.5em}
\subsection{Proof search}
\label{sec:pahse2}
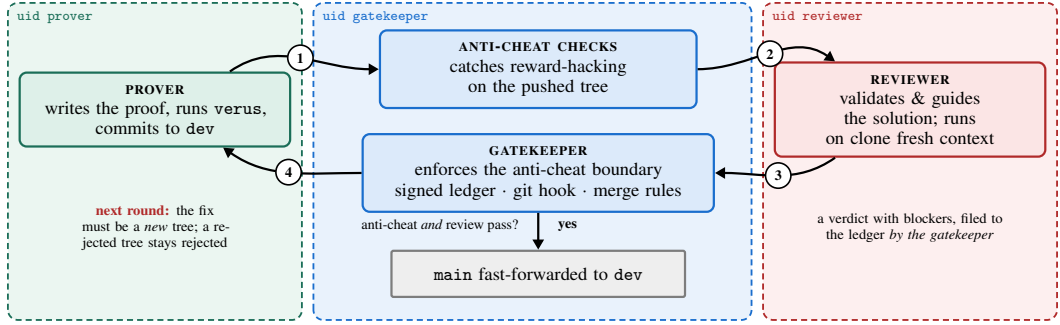
\begin{figure}[t]
  \centering
  \resizebox{\textwidth}{!}{% Figure: one round of the merge gate, read clockwise.  The three regions are
% the three Unix accounts; every step crosses an account boundary.  The merge
% condition and the fast-forward both hang off the gatekeeper's own box,
% because it is the gatekeeper's account that evaluates the ledger and moves
% the branch -- the prover can only ask.
%
% The three regions are drawn to the same height on purpose.  Green and red
% carry a little slack at the bottom as a result; that is cheaper than three
% ragged rectangles.
%
% Wide and shallow: once \resizebox scales the picture to \textwidth, the page
% height it costs is set by its aspect ratio, so width is nearly free.
%
% Three things to keep if you move anything:
%   * arrow endpoints are anchored to a node border, never to a hand-computed
%     coordinate -- a box is as tall as its text, not as tall as its
%     minimum height, so literal coordinates end up inside the box;
%   * a step node must sit *inside* the `to' operation --- a node written after
%     the target coordinate attaches to the coordinate, not to the path;
%   * every floating label carries fill=<its band colour>, so it masks the line
%     it sits on.  A label must therefore stay wholly inside one band.
\begin{tikzpicture}

  % ---- the three UID regions, all the same height -------------------------
  \begin{pgfonlayer}{tlback}
    \fill[tlband, fill=tlgreenband, draw=tlgreen] (-8.8,-2.90) rectangle (-3.7,2.6);
    \fill[tlband, fill=tlblueband,  draw=tlblue ] (-3.5,-2.90) rectangle ( 4.1,2.6);
    \fill[tlband, fill=tlredband,   draw=tlred  ] ( 4.3,-2.90) rectangle ( 9.4,2.6);
  \end{pgfonlayer}
  \node[tlbandlabel, text=tlgreen, fill=tlgreenband, anchor=north west]
        at (-8.75,2.56) {uid \textbf{prover}};
  \node[tlbandlabel, text=tlblue,  fill=tlblueband,  anchor=north west]
        at (-3.45,2.56) {uid \textbf{gatekeeper}};
  \node[tlbandlabel, text=tlred,   fill=tlredband,   anchor=north west]
        at ( 4.35,2.56) {uid \textbf{reviewer}};

  % ---- the four stations of the round -------------------------------------
  \node[tlgreenbox, text width=42mm, minimum height=11mm] (P) at (-6.25,0.75)
        {\tlrole{prover}\\writes the proof, runs \texttt{verus},\\
         commits to \texttt{dev}};

  \node[tlbluebox, text width=50mm, minimum height=11mm] (MECH) at (0.4,1.45)
        {\tlrole{anti-cheat checks}\\catches reward-hacking\\on the pushed tree};

  \node[tlredbox, text width=42mm, minimum height=11mm] (REV) at (6.85,0.75)
        {\tlrole{reviewer}\\validates \& guides the solution; runs\\on clone fresh context};

  \node[tlbluebox, text width=56mm, minimum height=11mm] (LED) at (0.4,-0.35)
        {\tlrole{gatekeeper}\\enforces the anti-cheat boundary\\signed ledger $\cdot$ git hook $\cdot$ merge rules};

  \node[tlgreybox, tlsquare, text width=46mm, minimum height=8mm] (MAIN)
        at (0.4,-2.10) {\texttt{main} fast-forwarded to \texttt{dev}};

  % ---- the round: four steps, clockwise -----------------------------------
  \draw[tlflow] ($(P.north)!0.55!(P.north east)$) to[out=35, in=180]
        node[tlstep, pos=0.5] {1} (MECH.west);

  \draw[tlflow] (MECH.east) to[out=0, in=145]
        node[tlstep, pos=0.5] {2} ($(REV.north)!0.55!(REV.north west)$);

  \draw[tlflow] ($(REV.south)!0.55!(REV.south west)$) to[out=-145, in=0]
        node[tlstep, pos=0.5] {3} (LED.east);

  \draw[tlflow] (LED.west) to[out=180, in=-35]
        node[tlstep, pos=0.5] {4} ($(P.south)!0.5!(P.south east)$);
  \node[tledge, fill=tlgreenband, text width=34mm] at (-6.25,-1.30)
        {\textbf{\textcolor{tlred}{next round:}} the fix must be a
         \emph{new} tree; a rejected tree stays rejected};

  \node[tledge, fill=tlredband, text width=34mm] at (6.85,-1.30)
        {a verdict with blockers, filed to\\the ledger \emph{by the gatekeeper}};

  % ---- the merge, which the gatekeeper decides and performs ---------------
  \draw[tlflow] (LED.south) -- (MAIN.north);
  \node[tledge, fill=tlblueband, anchor=east] at (0.15,-1.27)
        {anti-cheat \emph{and} review pass?};
  \node[tledge, fill=tlblueband, anchor=west] at (0.65,-1.27) {\textbf{yes}};
\end{tikzpicture}}
  \caption{One cycle of the merge gate. The prover and the reviewer are LLM agents; the
  gatekeeper is a set of scripts that orchestrate the pipeline. Three entities run as three Unix accounts.  Only the gatekeeper's account can merge: it does so at the prover's request, and only once the anti-cheat checks and the reviewer have both accepted.}
  \label{fig:uid-loop}
\end{figure}

\vspace{-0.5em}
%% GENERATED from docs/paper-section-prover-reviewer.tex by split-section.py -- edit THAT file.
We propose a prover--reviewer setup for automatic proof generation on Rust/Verus
systems (Figure~\ref{fig:uid-loop}).

Our setup has three components. 

\begin{enumerate}[leftmargin=1.5em, itemsep=2pt]
\vspace{-0.5em}
\item \textbf{The prover.} A long-running LLM session that writes the proof files, running
Verus locally for quick feedback and committing as it goes. 
\vspace{-0.2em}
\item \textbf{The reviewer.} An LLM session with clean context, called on request by the prover. The reviewer reads an incremental addition from the prover 
and returns a verdict with a
list of blockers.
\vspace{-0.2em}
\item \textbf{Anti-cheating checks.} We ensure that the prover cannot edit system specification 
or insert statements (e.g., \texttt{assume(false)}) that would make the proof vacuously true, passing the Verus verifier. We extend lynette~\cite{autoverus} with a wider class of anti-cheating checks, see Appendix~\ref{app:anticheat} for details.

\end{enumerate}
\vspace{-1em}
\paragraph{Motivation for the reviewer.}
The reviewer allows the prover to seek feedback incrementally. In long-horizon tasks, 
such as in complex systems, this means the prover can break down the proof into smaller pieces and change course early if it chooses
an unfruitful proof strategy.
Moreover, the Verus verifier is limited: it only shows which
assertion or postcondition failed, but no counterexample. 
The reviewer provides semantic feedback and can point out why and how to fix the solution. 
\vspace{-1em}
\paragraph{Technical setup.}
The two agents work in a local git repository, initialised with a Verus bundle of the
system and one property to prove about it. The bundle also contains a supporting temporal layer: the Verus encoding of temporal logic from
Anvil~\cite{sun2024anvil}, for the prover to use out-of-the-box. The
prover works incrementally on a \code{dev} branch, proving subsets of lemmas. To merge \code{dev} into \code{main}, we want to ensure the proof
is sound: all anti-cheat checks pass, and the reviewer has accepted the head
commit on \code{dev}. If the prover could merge into \code{main} on its own, it could breach these criteria. A set of programs we call the gatekeeper creates a deterministic boundary between the agents and the code. We
describe how this is enforced in Appendix~\ref{app:boundary}. One cycle of
operation is shown in Figure~\ref{fig:uid-loop}.

\vspace{-1em}
\paragraph{Review.}
The reviewer is a
second agentic LLM session that starts with a fresh context. The prover requests a review periodically. It has two main roles:
\vspace{-0.5em}
\begin{description}
\item \textbf{Validate or reject unprovability claims.} The prover may declare that a
  property is false. We use this escape hatch to identify errors in
  specifications, but the prover can misuse it to escape from a hard proof. This
  is why the reviewer is told to push back hard, retrace the counterexample on
  its own, and pragmatically stress-test the unprovability argument. In Appendix~\ref{app:refuted-disproof}, we show a run where the reviewer refused an unprovability claim from the
  prover, and the prover went on
  to prove the theorem. This is
significant for formal verification to be used in production: false alarms waste time in repairing a system that wasn't broken to start with.
\vspace{-0.5em}
\item \textbf{Guide the proof.} Steer the prover towards a cleaner proof. For example,
  instead of copy-pasting lemmas for mirror cases (left/right), the prover is
  advised to use a parameterised interface. This can lead to shorter and more
  human-maintainable proof code.
\end{description}

After the mechanical checks and the reviewer have run, the prover either requests a merge into main if they both accept, or continues iterating.

% ---------------------------------------------------------------------------

\vspace{-0.5em}
\subsection{From translations to an evaluation set}
\label{sec:dataset}

%\begin{itemize}
%    \item Flattened the file structured TLA+ tanslations produced by the LLM. The algorithmic one already gave the flattened one
%    \item Region1: frozen scaffold. Region2: task itsefl (spec). Region3: Added a proof function of the theorem needing proof with an empty body
%    \item tlaverus by llm dataset: 12352, tlaverus by algo dataset: 5160 (\cref{tab:translation-funnel})
%    \item Proven algo: 3094, llm: 492 
%    \item How we choose the 40 tasks for the eval dataset. 
%\end{itemize}
\vspace{-0.5em}
\paragraph{A single file with three regions.} After the two pipelines produced the algorithmic and agentic translations, both are normalised to the same three-region layout so models see identical input structure.
Region 1 is a frozen scaffold of temporal-logic definitions and proof rules—755 lines, 30,780 characters, byte-identical across tasks and never edited. Region 2 contains the problem: the state type, state machine, property predicates, and theorem statement. Region 3 is an \textit{empty} \verb|proof fn| that the model must fill.

\vspace{-0.5em}
\paragraph{Corpus.}
We ran both pipelines over the same 16,459 (specification, property) pairs; \Cref{tab:translation-funnel} summarizes the results. A reference proof, where available, must verify without errors and be a pure addition to the task, identical in regions 1 and 2, so it cannot restate the problem. Since well-posedness only means the verifier can parse the problem, we also ran proof search. At scale, the algorithmic pipeline found proofs for \textbf{3,094} out of 5,160. For the agentic pipeline, from the 12,352 well-posed translations, we ran a 500-task sample drawn from problems already proved algorithmically, obtaining \textbf{492} proofs. This enables a paired comparison: both encodings are provable, so failures reflect the encoding rather than an unprovable goal. Dataset statistics are given in \Cref{app:dataset-stats}: task counts by property type over all
well-posed translations (\Cref{app:stats-wellposed}), and counts and mean task lengths over the
proven subset (\Cref{app:stats-proven}).

\vspace{-1em}
\paragraph{Selecting the 40 evaluation tasks.}
The evaluation set is 40 tasks, 10 from each of the four property types (\emph{safety},
\texttt{TypeOK}, \emph{invariance}, \emph{liveness} -- see Table~\ref{tab:property-types}), each present in both encodings. Candidates were
filtered on seven criteria: (1) both translations
exist; (2) the task is known provable; (3) the two translations
state the same theorem; (4) the source repository carries a permissive licence; (5) the specification is not from a TLA$^+$ tutorial or benchmark repository, which a
model has likely seen in training;
 (6) no near-duplicate
specifications; and (7) specification and repository diversity,
applied when sampling rather than as an exclusion.
\vspace{-0.5em}

\vspace{0.5em}
\begin{center}
\small
\begin{tabular}{@{}lll@{}}
\toprule
\textbf{Category} & \textbf{Form} & \textbf{Meaning} \\
\midrule
\texttt{safety}          & $\Box[A]_{\mathit{vars}}$, $\Box(P \Rightarrow \Box Q)$, and similar & nothing bad ever happens \\
\texttt{safety\_typeOK}  & $\Box\,\mathit{TypeOK}$ & the variables stay of the right type \\
\texttt{invariance}      & $\Box P$, $P$ a state predicate other than $\mathit{TypeOK}$ & $P$ holds in every reachable state \\
\texttt{liveness}        & $P \leadsto Q$ or $\Diamond P$ & something good eventually happens \\
\bottomrule
\end{tabular}
\captionof{table}{The four property categories every (specification, property) pair is classified into.}
\label{tab:property-types}
\end{center}
\vspace{-1em}

\begin{table}[t]
\centering

\begin{subtable}[t]{0.48\textwidth}
\centering
\begin{tabular}{lr}
\toprule
& count \\
\midrule
Pairs attempted & 16,459 \\
Not translated & 10,389 \\
Translated & 6,070 \\
\quad$\rightarrow$~trivial / vacuous & 145 \\
\quad$\rightarrow$~Verus rejected output & 765 \\
\quad$\rightarrow$~well-posed translations & 5,160 \\
\qquad$\rightarrow$~\textbf{proven} & \textbf{3094} \\
\bottomrule
\end{tabular}
\caption{Algorithmic (\texttt{TranSpec})}
\label{tab:funnel-algo}
\end{subtable}
\hfill
\begin{subtable}[t]{0.48\textwidth}
\centering
\begin{tabular}{lr}
\toprule
& count \\
\midrule
Pairs attempted & 16,459 \\
Judged unprovable & 2,851 \\
Judged provable & 13,608 \\
\quad$\rightarrow$~trivial / vacuous & 1,235 \\
\quad$\rightarrow$~property type: unknown & 21 \\
\quad$\rightarrow$~well-posed translations & 12,352 \\
\qquad$\rightarrow$~\textbf{proven} (of 500 sampled) & \textbf{492} \\
\bottomrule
\end{tabular}
\caption{Agentic (LLM)}
\label{tab:funnel-agentic}
\end{subtable}

\caption{Translating the same 16,459 TLA$^+$ (spec, property) pairs to Verus, as a funnel.
A \emph{well-posed} translations verifies at exactly one error---the open proof obligation;
a \emph{trivial} one already verifies with an empty stub.
The pipelines lose tasks at different stages: the algorithmic one cannot translate them,
the agentic one judges them false. All the 'Judged provable' tasks verifies by Verus.}
\label{tab:translation-funnel}
\vspace{-1.5em}
\end{table}

\subsection{Experimental setup}
\label{sec:setup}

\begin{wraptable}{r}{0.46\textwidth}
\centering
\small
\begin{tabular}{lrr}
\toprule
Dataset & Tasks & Mean chars \\ %& Mean lines \\
\midrule
\texttt{verusbench}     & 150 & 619\\% & 33 \\
\texttt{tlaverus\_algo} & 40 & 47,379 \\ %& 1,034 \\
\texttt{tlaverus\_llm}  & 40 & 54,800 \\ %& 1,391 \\
\bottomrule
\end{tabular}
\caption{\textbf{Size of the three evaluated datasets.} Both include the frozen scaffold of 30,780
characters shared by every \texttt{tlaverus} task; net of it the means are 16,599 and 24,020.}
\vspace{-1em}
\label{tab:dataset-size}
\end{wraptable}

\vspace{-0.5em}
\paragraph{Datasets.} We evaluate off-the-shelf large language models on three datasets: \texttt{tlaverus\_algo} and
\texttt{tlaverus\_llm}, the two encodings of the 40 paired tasks built in
\Cref{sec:dataset}, and \texttt{verusbench} (\cite{autoverus}), an existing benchmark of 150 short, self-contained
Verus functions. \Cref{tab:dataset-size} gives their sizes. The contrast is the point of including
\texttt{verusbench}: a \texttt{tlaverus} task is roughly $80\times$ longer, because the model must
prove a property inside a whole translated specification rather than about a single function. The two
\texttt{tlaverus} rows are the same 40 \mbox{(spec, property)} pairs translated two ways, so their difference, 7,421 characters,
is the cost of the translation style alone.  \Cref{tab:dataset-size-bytype} reports the
same 40 tasks net of the 30{,}780-character scaffold both task shares, leaving the
problem text alone, and shows that the gap is not concentrated in one property type: it holds at
$1.35$--$1.55\times$ across all four.
\vspace{-0.5em}
\paragraph{Evaluation.} We evaluate eleven models, from frontier closed-weight systems down to
open-weight models of 35B--120B parameters; \Cref{app:setup} gives the full list with parameter
counts where published, decoding parameters, toolchain versions and more.\\
The evaluation harness differs from the prover--reviewer setup. Every \texttt{tlaverus} task is
presented the same way: the model receives the task file and returns region~3 alone, the body of the
open \verb|proof fn|, and the harness re-attaches regions~1 and~2 before verification. This is cost
efficient, and a model cannot alter the problem by rewriting the file around it. \texttt{verusbench}
tasks are much shorter, have no separate regions, and the model returns the whole file. Verus is the
oracle: a submission passes only if the whole file verifies with zero errors. Otherwise the
verifier's error output is fed back for revision, up to $V=10$ verification attempts per sample. We
draw $k=3$ independent samples per task and report pass@$k$, the fraction of tasks solved by at least
one sample, alongside the per-sample pass rate. Passing files are additionally checked for cheating
(\Cref{sec:pahse2}, \Cref{app:anticheat}).

\begin{wraptable}{r}{0.46\textwidth}
\vspace{-\intextsep}
\centering
\small
\begin{tabular}{lrrr}
\toprule
Property type & \texttt{algo} & \texttt{agent} & Ratio \\
\midrule
safety          & 18{,}113 & 24{,}460 & 1.35$\times$ \\
\texttt{TypeOK} & 12{,}850 & 19{,}922 & 1.55$\times$ \\
invariance      & 17{,}707 & 25{,}800 & 1.46$\times$ \\
liveness        & 17{,}727 & 25{,}896 & 1.46$\times$ \\
\midrule
All             & 16{,}599 & 24{,}020 & 1.45$\times$ \\
\bottomrule
\end{tabular}
\caption{\textbf{Problem text only, net of scaffold.} Mean length of the 40 paired tasks in
characters, with the 30{,}780-character scaffold subtracted, 10 tasks per property type.}
\label{tab:dataset-size-bytype}
\vspace{-2em}
\end{wraptable}
\vspace{-0.5em}

\section{Results}
\label{sec:eval}
\vspace{-1em}
In this section, we compare proof search across the two encodings by running the prover–reviewer pipeline
on the same 500 (specification, property) pairs. Moreover, we evaluate a range of state-of-the-art language models on three datasets.
\vspace{-0.5em}
\subsection{Agentic vs algorithmic translation comparison}
\vspace{-1em}
\label{sec:agentvsalg}
%% Section 4.1 -- Agentic vs algorithmic translation. Draft 2026-09-14.
%% Figures: the Appendix G fragments restyled as fig2_forest_sec41 / fig10_verus_run_outcomes_sec41; their labels live here now.
\providecommand{\draft}[1]{\begin{draftblock}#1\end{draftblock}}

To compare the two translations we ran the same prover on the 500 pairs twice: once from the \textsc{transpec} translation and once from the agentic one. The 500 were drawn from pairs already proven algorithmically, so every algorithmic side is proven by construction; on the agentic side the prover closed 492 and exhausted its budget on 8. All paired statistics below are over the 492 pairs proven in both encodings.

\vspace{-1em}
\paragraph{The LLM translation costs 19\% more to prove.}
It is the more expensive one in 71\% of pairs, and the median ratio is $1.19\times$ $[1.15, 1.23]$:
\$2.07 against \$1.67 per proof (\Cref{fig:llm-alg-cost}). Proofs are the same length ($1.05\times$), and Z3 does less work on the LLM side, $0.86\times$
the rlimit and $0.73\times$ the quantifier instantiations (\Cref{fig:llm-alg-rlimit}). The extra API cost goes into
turns: $1.38\times$ prover turns, $2.00\times$ verus runs and $1.88\times$ input tokens per proof
(\Cref{fig:llm-alg-tokens}). 
\vspace{-1em}
\paragraph{The extra verus runs are mostly compile errors.}
\Cref{fig:llm-alg-verus-runs} splits the runs by outcome. Both sides need 1.62 successful runs per
proof, and the LLM side has fewer failed verifications (0.32 against 0.64). The main difference is
trigger errors, 1.19 runs per proof against 0.13, and name-resolution errors, 0.26 against 0.07.
Two thirds of LLM proofs hit at least one trigger error; only 8\% of algorithmic proofs do.
\vspace{-1em}
\paragraph{The algorithmic translation leaves more triggers to Verus, yet causes fewer trigger errors.}
\textsc{transpec} emits three times as many \texttt{forall} quantifiers as the LLM translator and
annotates fewer of them: in 47\% of algorithmic bundles Verus had to pick triggers for six or more
quantifiers, while 80\% of LLM bundles have none (\Cref{fig:llm-alg-triggers}).
\vspace{-0.5em}

\subsection{Model evaluations}
\label{sec:modeleval}

\begin{figure}[H]
\centering
\includegraphics[width=\linewidth]{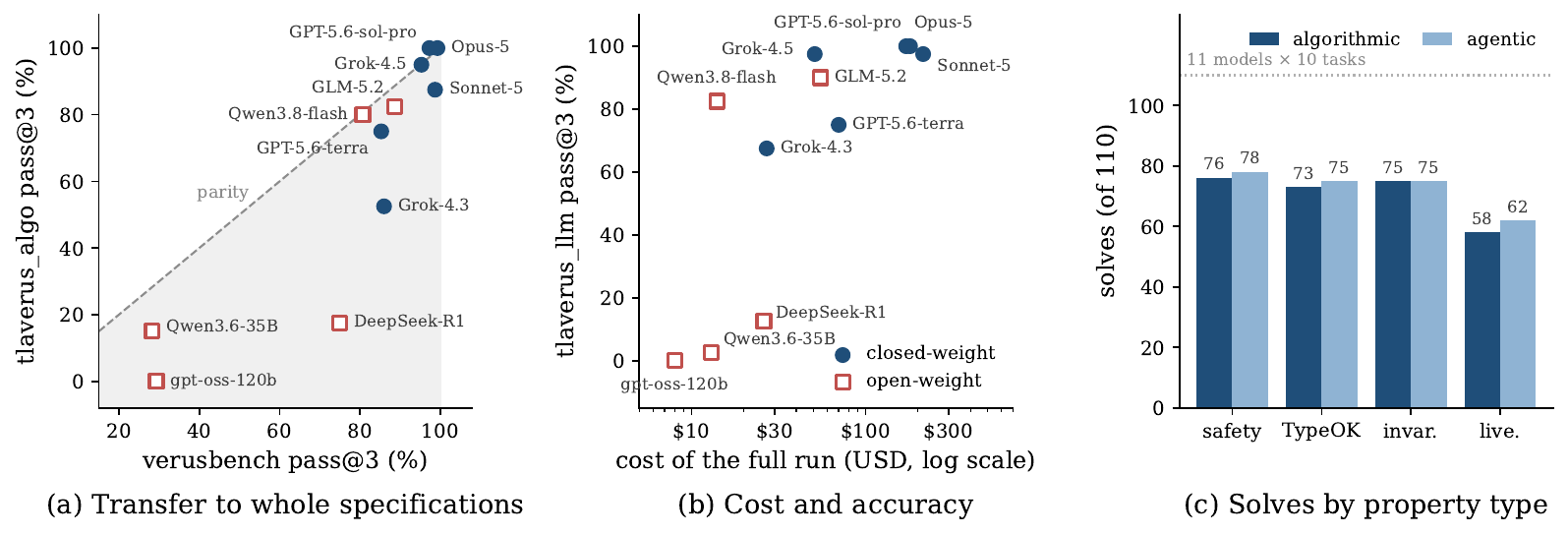}
\caption{\textbf{(a)} \texttt{verusbench} against \texttt{tlaverus\_algo} pass@3; dashed line is
parity, shading is where a model does worse on \texttt{tlaverus}. \textbf{(b)} Cost of the full \texttt{tlaverus\_llm}
run against pass@3. \textbf{(c)} Tasks solved by property type, summed over all 11 models: each bar is out of 110 (11 models $\times$ 10 tasks of that type).}
\label{fig:results-overview}
\vspace{-0.5em}
\end{figure}

We evaluate eleven models on all three datasets. \Cref{tab:full-vb} to \Cref{tab:by-type-model} in \Cref{app:eval_results} shows all numbers used for creating \Cref{fig:results-overview} and \Cref{tab:model-results}. The main measure of accuracy is pass@$3$. 
\vspace{-0.5em}

\paragraph{TLA-Verus is harder than Verusbench for every model but the strongest ones.}
Nine of the eleven models score lower on
\texttt{tlaverus\_algo}, and only Opus-5 and GPT-5.6-sol-pro, the best ones gain; Grok-4.5 and Qwen3.8-flash are the boundary cases at $-0.3$ and $-0.7$. The rest lose heavily: 33.5 points for Grok-4.3, 57.2 for DeepSeek-R1, and 29.3 for
gpt-oss-120b, which reaches zero (\Cref{fig:results-overview}a). %On \texttt{tlaverus\_llm} the loss is less widespread, six of eleven falling (\Cref{tab:model-results}). The two gainers simply saturate, solving 40 of 40 on both datasets. Below it the ordering changes: a short-function benchmark does not predict the ordering on whole specifications.

\vspace{-0.5em}
\paragraph{Tlaverus\_algo is not easier than tlaverus\_llm despite consisting shorter tasks.} The two datsets have the same 40 pairs, so this comparison is controlled. For nine of eleven models, the agentic translation is solved as well as or better than the algorithmic translation, despite the tasks in \texttt{tlaverus\_llm} being $1.45\times$ longer on average. (\Cref{tab:dataset-size-bytype}), the opposite of what length alone would predict. The only two models which lose on \texttt{tlaverus\_llm} (DeepSeek-R1 and Qwen3.6-35B) are among the weakest. The ordering of the models remains the same for two \texttt{TLA-Verus} datasets (\Cref{tab:model-results}), with rank changes confined to ties. We hypothesize that \texttt{tlaverus\_llm} is easier for the models, because the agentic translation of the tasks may contain additional comments helping in proof generation.
\vspace{-0.5em}
%\begin{wrapfigure}{r}{0.42\textwidth}
%\vspace{-\intextsep}
%\centering
%\includegraphics[width=0.4\textwidth]{figures/fig-by-type.pdf}
%\caption{\textbf{Solves by property type}, over all eleven models, out of 110 model--task pairs per
%type and half.}
%\label{fig:by-type}
%\vspace{-1em}
%\end{wrapfigure}
\paragraph{Liveness is the hardest type, and not because these tasks are the longest.}
Summed over all eleven models, safety, \texttt{TypeOK} and invariance are solved 76, 73 and 75 times
out of 110 on \texttt{tlaverus\_algo}, while liveness reaches only 58 (\Cref{fig:results-overview}c). \texttt{tlaverus\_llm} behaves the same way: 78, 75 and 75 against 62 for liveness. Liveness is where strong models break first: Sonnet-5
holds 10/10 on safety and invariance on \texttt{tlaverus\_algo} and drops to 7/10 on liveness,
GPT-5.6-terra goes 10/10 to 4/10. Length does not explain it --- safety and liveness differ by under
2\% in problem-text length (\Cref{tab:dataset-size-bytype}), and \texttt{TypeOK}, the shortest type,
is not the easiest.
\vspace{-0.5em}
\paragraph{Three attempts reach well beyond one.} Opus-5 is the only model that solves tasks with no gap on either dataset. gpt-oss-120b has no gap on \texttt{tlaverus\_algo} and \texttt{tlaverus\_llm} since it does not solve anything.
For every other model, pass@3 exceeds the per-sample rate, and often by a lot (\Cref{tab:model-results}): e.g. GPT-5.6-sol-pro
solves every task on \texttt{tlaverus\_llm} but passes only 83.3\% of samples, and GPT-5.6-terra
reads 75.0 against 52.5. The two numbers answer different questions --- what a model can reach when
allowed to retry, and how often a single attempt succeeds --- so we report both throughout.
\vspace{-0.5em}

\paragraph{The weakest models spend the budget without getting closer.}\looseness=-1
Every model below 20\% uses all ten verification attempts on nearly every sample and passes almost
none of them. 
Turns track strength almost perfectly in the other direction too:
Opus-5 needs only 1.1 per sample on \texttt{tlaverus\_algo}, and the count climbs down the table
until the budget is exhausted (\Cref{tab:model-results}; failure analysis in \Cref{app:failure-analysis}).
\vspace{-0.5em}

\paragraph{Price is an unreliable guide to accuracy, and reasoning volume is no guide at all.}
Stronger models are generally the more expensive ones on both \texttt{tlaverus} datasets
(\Cref{fig:results-overview}b, \Cref{fig:cost-reasoning}a). The interesting cases are the models that sit well off that line, and
they are the same on both datasets. Grok-4.5 is the clearest: it reaches 97.5\% on
\texttt{tlaverus\_llm} for \$51, the same score Sonnet-5 reaches for \$214, at a quarter of the price.
Qwen3.8-flash is among the cheapest at \$14 and still lands mid-field at 82.5\%, ahead of GPT-5.6-terra
at \$70.
Reasoning volume, meanwhile, carries no
signal at all (\Cref{fig:cost-reasoning}b,~c). 

\begin{table*}[t]
\centering
\small
\setlength{\tabcolsep}{4.5pt}
\begin{tabular}{lrrrrrrrrrrr}
\toprule
& \multicolumn{2}{c}{\texttt{verusbench}} & \multicolumn{4}{c}{\texttt{tlaverus\_algo}} & \multicolumn{4}{c}{\texttt{tlaverus\_llm}} & \\
\cmidrule(lr){2-3}\cmidrule(lr){4-7}\cmidrule(lr){8-11}
Model & p@3 & p@1 & p@3$\downarrow$ & p@1 & turns & reas. & p@3$\downarrow$ & p@1 & turns & reas. & $\Delta (\%)$ \\
\midrule
Opus-5                  &  99.3 &  99.3 & \textbf{100.0} & 100.0 &  1.1 &  36.0k & \textbf{100.0} & 100.0 &  1.6 &  38.3k &     0.0 \\
GPT-5.6-sol-pro         &  97.3 &  93.3 & \textbf{100.0} &  92.5 &  3.5 &  16.8k & \textbf{100.0} &  83.3 &  5.5 &  18.2k &     0.0 \\
Grok-4.5                &  95.3 &  90.4 &  95.0 &  90.8 &  4.2 &   6.6k &  97.5 &  90.0 &  5.0 &   4.6k &  $+2.5$ \\
Sonnet-5                &  98.7 &  94.2 &  87.5 &  82.5 &  3.4 & 196.9k &  97.5 &  87.5 &  4.6 & 115.2k & $+10.0$ \\
GLM-5.2$^{\circ}$       &  88.7 &  75.6 &  82.5 &  70.8 &  5.7 & 101.6k &  90.0 &  78.3 &  5.6 &  72.5k &  $+7.5$ \\
Qwen3.8-flash &   80.7 &   58.2 &  80.0 &  68.3 &  6.4 & 217.9k &  82.5 &  60.0 &  7.3 & 169.6k & $+2.5$ \\
GPT-5.6-terra           &  85.3 &  73.1 &  75.0 &  60.8 &  6.8 &   5.9k &  75.0 &  52.5 &  7.5 &   4.7k &     0.0 \\
Grok-4.3                &  86.0 &  78.2 &  52.5 &  40.8 &  8.0 &  46.7k &  67.5 &  52.5 &  7.6 &  36.7k & $+15.0$ \\
DeepSeek-R1$^{\circ}$ & 74.7 & 57.6 & 17.5 &  10.8 &  9.6 &  32.4k &  12.5 &   7.5 &  9.8 &  29.8k &  $-5.0$ \\
Qwen3.6-35B$^{\circ}$   &  28 &  14.4 &  15.0 &   5.0 &  9.9 &  64.3k &   2.5 &   0.8 & 10.0 &  69.0k & $-12.5$ \\
gpt-oss-120b$^{\circ}$  &  29.3 &  13.1 &   0.0 &   0.0 & 10.0 &  26.6k &   0.0 &   0.0 & 10.0 &  27.2k &     0.0 \\
%\midrule
%Mean$^{\ast}$           &  78.3 &  69.0 &  62.5 &  55.4 &  6.2 &  59.7k &  64.2 &  55.2 &  6.7 &  48.2k &  $+1.8$ \\
\bottomrule
\end{tabular}
\vspace{0.5em}
\caption{\textbf{Model results.} \emph{p@3} and \emph{p@1} (\%): tasks solved by at least one of $k=3$
samples out of 40, and all samples passed (120). \emph{turns} and \emph{reas.}: mean model calls and reasoning tokens per
sample. $\Delta$: \texttt{tlaverus\_llm} minus \texttt{tlaverus\_algo} p@3.
$^{\circ}$~open-weight. Remaining cols in \Cref{tab:full-vb,tab:full-algo,tab:full-llm}. } %Models ordered by p@3 \texttt{tlaverus\_algo} and \texttt{tlaverus\_llm}. 
\label{tab:model-results}
\vspace{-1em}
\end{table*}

%---------------------------
\vspace{-0.5em}
\section{Related works}
\vspace{-0.5em}
\paragraph{TLA$^+$, its tooling, and the gap to code.}
TLA$^+$ \citep{lamport1994tla,lamport2002specifying} is written by engineers, not only by
verification researchers: AWS reports using it on S3, DynamoDB and EBS \citep{newcombe2015aws}, and
MongoDB on replication and reconfiguration \citep{schultz2022reconfig}. TLC \citep{yu1999tlc} and
Apalache \citep{konnov2019apalache} check bounded instances and return counterexamples, not proofs;
TLAPS proves and is effective on safety \citep{cousineau2012tlaps,chaudhuri2010safety}, but liveness
must be reduced to a propositional fragment and system-scale temporal proofs stay rare. Nor is there
a refinement path to an implementation: PGo compiles Modular PlusCal to Go \citep{hackett2023pgo},
trace validation checks a running system against its specification \citep{cirstea2024trace}, and
IronFleet \citep{hawblitzel2015ironfleet} and Verdi \citep{wilcox2015verdi} close the gap by
construction, at the price of writing the system inside the framework. We keep the specification and
change the prover, moving existing TLA$^+$ into Verus \citep{lattuada2023verus,lattuada2024verus},
an auto-active Rust verifier, so that specification and implementation can share a language and liveness
is discharged by the same deduction as safety, at a scale hand-written TLAPS proofs have not reached.
\vspace{-1em}

\paragraph{Liveness proofs.}
Temporal reasoning is not native to auto-active verifiers but can be embedded: Anvil defines a
TLA-style temporal logic as a Verus library and proves liveness of Kubernetes controllers
\citep{sun2024anvil}, and our encoding builds on it. In the first-order setting, Ivy
\citep{padon2016ivy} and liveness-to-safety make liveness decidable in a restricted fragment, at the
cost of manual prophecy and fairness encodings \citep{padon2018liveness,padon2017paxos}. Each such
proof is a bespoke expert artifact; we generate them automatically over a corpus nobody wrote proofs
for.
\vspace{-1em}

\paragraph{LLMs for Verus.}
AutoVerus orchestrates agents that mimic the phases of human proof construction \citep{autoverus};
SAFE and VeruSyn bootstrap Verus-capable models by self-evolution, the latter scaling to 6.9M
synthesised verified programs \citep{chen2024safe,di2026verusyn}, on which supervised fine-tuning
lifts a mid-size open-weight model to near-parity with a $\sim$50$\times$ larger frontier model
\citep{ujvary2026lightning}; RagVerus retrieves same-project proofs \citep{zhong2025ragverus}; and
AlphaVerus translates Dafny to Verus \citep{aggarwal2024alphaverus}, where ours is formal-to-formal
and accepted only when Verus compiles the bundle to exactly one open obligation. All target
functional correctness of Rust code, whereas our obligations are temporal claims about infinite
behaviours. They also share our scaffold shape --- verifier errors fed back to the model, with cheat
detection on top: VeruSAGE runs a cheat-checker \citep{yang2025verusage} and AlphaVerus pairs a
rule-based detector with a filtering model. We keep both parts (\cref{sec:pahse2},
\cref{app:anticheat}), run under a different user than the prover so it cannot write its own verdict,
and record every verdict in a signed ledger.
\vspace{-1em}

\paragraph{Verification benchmarks.}
Verus benchmarks have grown from 150 self-contained functions in \texttt{verusbench}
\citep{autoverus} to 383 repository-level tasks in RepoVBench \citep{zhong2025ragverus} and 849
across eight projects in VeruSAGE-Bench \citep{yang2025verusage}, with comparable suites for Dafny
and Lean \citep{loughridge2024dafnybench,dougherty2025fvapps,ye2025verina,bursuc2025vericoding}.
Ours is generated rather than curated, and it is the first whose tasks are temporal --- including
liveness, whose fairness and well-foundedness arguments have no analogue in functional verification.
The specifications come from deployed systems and the Verus encodings did not exist before this
work, so the solutions are in no model's training data.

\vspace{-0.5em}
\section{Discussion}
\label{sec:discussion}
\vspace{-1em}
We have created a pipeline to translate TLA$^+$ system specifications to Verus, and create fully
machine-checkable proofs of safety and liveness properties. From over 16,000 (specification, property) pairs, our pipeline produces over 3,000 fully machine-checked proofs. Evaluating eleven models on a paired evaluation set shows that proof search on \texttt{tlaverus} is already saturated for the strongest frontier models -- Opus-5 and GPT-5.6-sol-pro both solve every task -- but the benchmark remains sharply discriminating further down the field, where scores span from 97.5\% (Grok-4.5) down to 0\% (gpt-oss-120b).

Our pipeline's limitation today is the inability to reliably differentiate between false statements and true statements for which finding a proof is too difficult. In some cases, properties require additional assumptions to be added to the Verus specification, and determining when an assumption is a \emph{fair} addition, similarly remains a challenge. Translations are not formally guaranteed to be faithful to their TLA$^+$ source. Towards faithfulness, \texttt{TranSpec} applies fixed, deterministic rules to each TLA$^+$ construct, agentic translations are checked by an LLM reviewer. Further, our model evaluation consists of a single run per model ($k{=}3$ samples on each of the 40 tasks), and we report pass@$k$ and per-sample pass rates as point estimates without uncertainty estimates. Porting temporal logic models to Verus opens the door to closing the implementation--specification gap itself: expanding the pipeline to produce fully end-to-end verified implementations of real systems, where safety and liveness proofs live alongside the implementation rather than in a separate TLA$^+$ model.
\vspace{-0.5em}

\newpage
\bibliographystyle{plainnat}   % natbib's own style; see make-main.py
\bibliography{references}   % no .bib extension
\newpage

%%%%%%%%%%%%%%%%%%%%%%%%%%%%%%%%%%%%%%%%%%%%%%%%%%%%%%%%%%%%

\appendix

% Appendix: anti-cheat checks. Short form; the long draft with the corpus measurement is
% docs/paper-section-anticheat-v2.tex. Numbers deliberately omitted: the gate is not final.

\section{Anti-cheat checks}
\label{app:anticheat}

The Verus verifier certifies that a proof is accepted, not that it is trustworthy. Like every
practical proof assistant, Verus provides \emph{escape hatches} that let an author declare a
fact instead of proving it. For instance, \code{assume(P)} adds $P$ as a hypothesis, and a
function marked \code{external\_body} has its \code{ensures} taken as true without its body
being checked. They exist to state trusted facts
about hardware, libraries and the environment, which lie outside the verified code. Used inside
a proof, they make it \emph{vacuous}: Verus reports \emph{N verified, 0 errors} while the proof
assumes the program is correct rather than proving it. An agent whose goal is to pass
verification can misuse them, so in our setup a proof counts as a solution only if it also passes
the two deterministic checks below, one for each way of making a task easier: changing the task,
or faking the proof.

\paragraph{The task is unchanged.}
lynette~\cite{autoverus} is the structural diff from AutoVerus: it parses the task file
and the candidate and requires that everything defining the problem statement is identical
(function signatures, \code{requires}/\code{ensures} clauses, \code{spec fn} bodies and
executable bodies -- with proof annotations removed). It allows only new proof or spec items to be added and nothing to be
deleted. A candidate that passes lynette has solved the original problem rather than a weaker
one.

\paragraph{The proof is not faked.}
lynette deliberately does not check proof bodies, so that the agent can prove lemmas any way it
likes. Escape hatches inside proof code, such as \code{proof \{ admit(); \}}, would therefore pass
it. We extend lynette with anti-cheating checks that reject any trust marker that is
\emph{net-new}, that is, appearing in the candidate but not in the task, ignoring
comments. Table~\ref{tab:anticheat} shows examples of cheating patterns that our checks flag.

% Needs only booktabs (in the main preamble). No `array' package: the cells are
% p{} columns with \raggedright inside and rows end in \tabularnewline, which the
% LaTeX kernel defines, so a pasted copy renders the same in every preamble.
\begin{table}[htbp]
\centering\footnotesize
\renewcommand{\arraystretch}{1.15}
\begin{tabular}{@{}p{0.15\linewidth}p{0.54\linewidth}p{0.25\linewidth}@{}}
\toprule
Cheat form & Example & Effect \tabularnewline
\midrule
\raggedright \code{admit} &
\raggedright \code{proof \{ admit(); \}}\newline \code{use vstd::pervasive::admit as shh;} &
\raggedright the current goal is accepted without proof \tabularnewline
\midrule
\raggedright \code{assume} &
\raggedright \code{assume(false)}\newline \code{macro\_rules! m \{ () => \{ assume(false) \} \}}\newline \hspace*{1.5em}\code{proof \{ m!(); \}} &
\raggedright the stated fact is accepted without proof \tabularnewline
\midrule
\raggedright \code{external\_body} &
\raggedright \code{\#[verifier::external\_body]}\newline \hspace*{1.5em}\code{proof fn any\_fact() ensures false \{\}} &
\raggedright the function's \code{ensures} is accepted without checking its body \tabularnewline
\midrule
\raggedright \code{axiom} &
\raggedright \code{axiom fn any\_fact() ensures false;} &
\raggedright the lemma is accepted without a proof \tabularnewline
\bottomrule
\end{tabular}
\vspace{0.6em}
\caption{Net-new escape hatches that our anti-cheating checks reject in the proof.}
\label{tab:anticheat}
\end{table}

%% GENERATED from docs/paper-section-prover-reviewer.tex by split-section.py -- edit THAT file.
\section{Enforcing the boundary}
\label{app:boundary}

Our requirement for merging \code{dev} into \code{main} is for both the
mechanical checks and the reviewer to accept. An autonomous prover is normally given \code{sudo}
permissions, which means it could merge into the repository directly, bypassing our
criteria for a sound proof. To mitigate this, our setup sandboxes the agents and implements safe 
communication between them. We explain four key components: Linux UIDs, MCP tools, a signed ledger, and git management.

\paragraph{MCP tools.}
Model Context Protocol (MCP) is a standard released by Anthropic for setting up
communication between AI models and tools. Our prover agent has access to five
tools:
\begin{enumerate}[leftmargin=1.5em, itemsep=1pt]
\item \code{run\_mechanical(sha?)} -- run mechanical checks on a commit, receiving the full report;
\item \code{run\_review(sha?)} -- request a review of a commit, receiving the full review;
\item \code{request\_merge(sha?)} -- request to fast-forward \code{main} to \code{dev}. The merge is allowed if the 
head commit carries an accepted review and mechanical checks; the head commit never carried rejected verdicts; the signed ledger is consistent, among other rules.
\item \code{merge\_status(sha?)} -- request the pass/fail status for every merge rule for the given commit, including reasons for failing each rule;
\item \code{push\_status()} -- report what commit is head for
  \code{dev} and \code{main} in the canonical repository, how many commits \code{dev} is ahead, and the
  most recent pushes the git hook refused.
\end{enumerate}
The
tools run under the prover's Linux UID and only give the request to the gatekeeper
through \code{sudo}. Using these tools, the prover can contribute to the git repo without having direct access to it. 

\paragraph{Git management.}
The prover's clone reaches the repository only over git's \texttt{ext}
transport. The remote's command is a root-owned helper that runs as the prover
with no privileges and accepts only the \texttt{upload-pack} and
\texttt{receive-pack} service names. It sends the request through \texttt{sudo}
to the gatekeeper's server, which runs the git service. The prover therefore has
no filesystem access to the repository and fetches and pushes through the
gatekeeper's uid. A root-owned \texttt{pre-receive} hook accepts pushes to
\texttt{dev}, refuses deletion of either branch and any other ref, and moves
\texttt{main} only by fast-forward to a commit that passes the same merge
decision function as \texttt{request\_merge()}.

\paragraph{Signed ledger.}
All mechanical checks, reviews, and merges are recorded in an append-only
ledger owned by the gatekeeper, signed with an Ed25519 key. The prover
is denied read of the key, as it is stored under a `deny to everyone else'
\code{0600} mode owned by the gatekeeper. It is also denied write of the ledger,
as that is a gatekeeper-owned directory under \code{0755} permission, and each
record is written once with \code{0444} permission. Each record contains the
hash of the previous one. Deleting a record would be detected by a check inside the merge decision function on each tool call. 

\paragraph{Linux UIDs.}
The prover, the gatekeeper and the reviewer are three unprivileged
accounts with no shared group. The canonical repository is gatekeeper-owned and
unreadable to the prover. The ledger directory is gatekeeper-owned and
world-readable, so the prover can audit it but cannot write it. The signing key
is gatekeeper-owned with mode \texttt{0600}. The gatekeeper's programs live in a
root-owned directory. The prover's \texttt{sudoers} entry lets it run, as the
gatekeeper and with a reset environment, exactly eight argument lists: the five
MCP tool back-ends and three transport verbs (\texttt{upload-pack},
\texttt{receive-pack}, \texttt{log-refusal}). The prover sends its data over
\texttt{stdin}. The reviewer has no \texttt{sudo} entry. This ensures the prover
cannot approve its own work or tamper with the ledger, but can deliver its work
to the repository.

\section{The reviewer in action}
\label{app:reviewer}

\subsection{Steering towards a shorter proof}
\label{app:steering}

% Provenance (not for publication):
%   s3://tlaverus-reasonable/outputs/dataset_outputs/opus-4.8-attempt2/provable/liveness/
%     d8bc5db96d5bf25e9927aa56195a8bb871d2f2c7-eventual_response_theorem/
%   reviews at seq 3, then seq 5.

The reviewer steers the prover towards a shorter, quicker solution. On a
liveness obligation, the prover had begun proving each step of a WF1 chain as
its own lemma.

\begin{reviewerquote}[reviewer, round 1 --- verbatim]
Per-step WF1 pattern is about to be replicated $\sim$15$\times$; require a
single parameterised \code{wf1\_link} helper now, before duplication compounds.
\end{reviewerquote}

\begin{reviewerquote}[reviewer, round 2 --- verbatim]
Resolved and done well. RM-008 delivers exactly this\ldots{} Do not re-raise.
\end{reviewerquote}

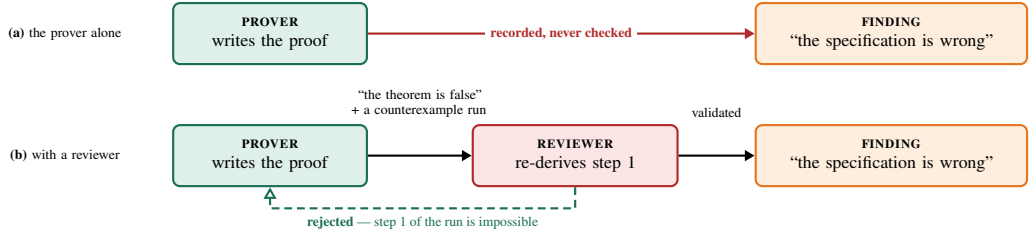
\begin{figure}[t]
  \centering
  \resizebox{\textwidth}{!}{% Figure: what the second agent adds, as two linear runs of the same claim.
%   (a) the prover alone: its unprovability claim becomes a finding unchecked.
%   (b) with a reviewer in the way: the claim is re-derived, and here refuted.
% The two rows share their columns, so the missing middle step in (a) is the
% point of the picture.
%
% Wide and shallow on purpose: page height follows the aspect ratio once the
% picture is scaled to \textwidth, so the row labels on the left are free.
\begin{tikzpicture}

  % ---- (a) the prover on its own ------------------------------------------
  \node[tlgreenbox,  text width=30mm, minimum height=11mm] (Pa) at (0,1.35)
        {\tlrole{prover}\\writes the proof};
  \node[tlorangebox, text width=44mm, minimum height=11mm] (Fa) at (11.2,1.35)
        {\tlrole{finding}\\``the specification is wrong''};
  \draw[tlflow, draw=tlred] (Pa) -- (Fa);
  \node[tledge, fill=white, text=tlred, font=\scriptsize\bfseries] at (5.25,1.35)
        {recorded, never checked};
  \node[tledge, anchor=east, text width=30mm] at (-2.10,1.35)
        {\textbf{(a)} the prover alone};

  % ---- (b) with a reviewer in the way -------------------------------------
  \node[tlgreenbox, text width=30mm, minimum height=11mm] (Pb) at (0,-0.85)
        {\tlrole{prover}\\writes the proof};
  \node[tlredbox,   text width=32mm, minimum height=11mm] (Rb) at (5.5,-0.85)
        {\tlrole{reviewer}\\re-derives step 1};
  \node[tlorangebox, text width=44mm, minimum height=11mm] (Fb) at (11.2,-0.85)
        {\tlrole{finding}\\``the specification is wrong''};
  \draw[tlflow] (Pb) -- (Rb);
  \draw[tlflow] (Rb) -- (Fb);
  \node[tledge, anchor=south, text width=34mm] at (2.70,-0.25)
        {``the theorem is false''\\+ a counterexample run};
  \node[tledge, anchor=south] at (8.05,-0.25) {validated};
  \node[tledge, anchor=east, text width=30mm] at (-2.10,-0.85)
        {\textbf{(b)} with a reviewer};

  % the rejection, which is what actually happened
  \draw[tlfeedback, draw=tlgreen]
        (Rb.south) -- (5.5,-1.80) -- (0,-1.80) -- (Pb.south);
  \node[tledge, anchor=north, text=tlgreen] at (2.75,-1.85)
        {\textbf{rejected} --- step 1 of the run is impossible};
\end{tikzpicture}}
  \caption{What the second agent adds. Without it, an unprovability claim is
  recorded unchecked.}
  \label{fig:double-vote}
\end{figure}

% Appendix case study: a refuted disproof.  Demonstrates the four callout box
% styles -- proverquote (green), reviewerquote (red), tlacode (sand) -- on the
% GreyConsensus Agreement run of 2026-09-05.
%
% Provenance (not for publication):
%   s3://tlaverus-reasonable/outputs/bundle-auto/
%     d92f8a8f6c23bde51564b35450953b7bc444c1e2/Agreement/8ed60c2/20260905T144235Z/
%   Quotes paraphrased from reviews/PR_1.md, comment_1.md, PR_2.md.

\subsection{A refuted disproof}
\label{app:refuted-disproof}

We used our pipeline to prove \emph{Agreement}, the central safety property of a
Raft-style consensus protocol taken from a public
repository.\footnote{From a public GitHub repository, since removed by its
owner. We quote only the operators needed here; the full specification is
available to reviewers on request.}
Agreement states
that no two nodes are leaders in the same election term.

\begin{tlacode}
Agreement == \A n1, n2 \in Nodes :
    (role[n1] = "Leader" /\ role[n2] = "Leader"
     /\ currentTerm[n1] = currentTerm[n2]) => n1 = n2
\end{tlacode}

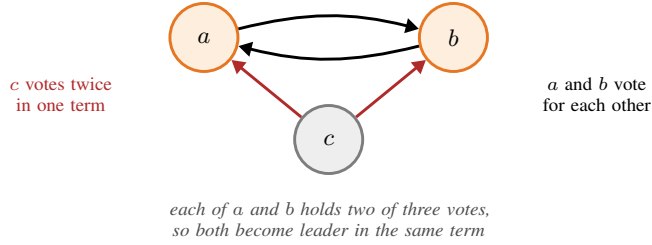
\begin{figure}[H]
  \centering
  % Figure: the prover's claimed counterexample to Agreement.  c grants a vote to
% two candidates in the same term, so a and b each hold two of three votes.
% (The reviewer showed the run is unreachable: StartPreVote is unsatisfiable for
% clusters of three or more, so no node can begin an election at all.)
%
% Small figure, ~7cm wide: use it at natural size, not \resizebox'd to
% \textwidth -- scaling it up would cost three times the page height.
% Leaders are the orange nodes; the caption says so, which is cheaper in
% vertical space than a badge over each one.
\begin{tikzpicture}
  \node[tlorangecircle, minimum size=9mm] (A) at (0,1.35)    {$a$};
  \node[tlorangecircle, minimum size=9mm] (B) at (3.3,1.35)  {$b$};
  \node[tlgreycircle,   minimum size=9mm] (C) at (1.65,0)    {$c$};

  % the two votes that should not both exist
  \draw[tlflow, draw=tlred] (C) -- (A);
  \draw[tlflow, draw=tlred] (C) -- (B);
  % and the two ordinary ones
  \draw[tlflow, bend left=14] (A) to (B);
  \draw[tlflow, bend left=14] (B) to (A);

  \node[tledge, text=tlred, anchor=east, text width=24mm] at (-0.60,0.60)
        {$c$ votes twice\\in one term};
  \node[tledge, anchor=west, text width=24mm]             at ( 3.90,0.60)
        {$a$ and $b$ vote\\for each other};
  \node[tlnote, text width=62mm] at (1.65,-1.05)
        {each of $a$ and $b$ holds two of three votes, so both become
         leader in the same term};
\end{tikzpicture}
  \caption{The prover's claimed counterexample to Agreement.}
  \label{fig:two-leaders}
\end{figure}

\noindent
The prover declined to prove and argued that the property was false. It supplied a
ten-step execution on a three-node cluster which resulted in two leaders in term~1. The prover
traced the error to the \texttt{AppendEntries} handler, which allows one node to
vote twice in a single term (Figure~\ref{fig:two-leaders}).

\begin{proverquote}[prover, iteration 1 --- paraphrased]
Agreement is not inductive and no strengthening rescues it: \texttt{votedFor}
is reset at equal term, so a node can grant two candidates in one term. Here is
a reachable execution in which $a$ and $b$ both become leader\ldots
\end{proverquote}

\noindent
The reviewer retraced the execution itself and found that step~1 is not a
transition of \texttt{Next}.

\begin{tlacode}
Send(m) == messages' = messages \cup {m}

StartPreVote(n) ==
    ...
    /\ \A m \in Nodes \ {n} : Send([type |-> "PreVote", ..., dest |-> m])
\end{tlacode}

\noindent
 \texttt{Send} requires \texttt{messages'} to equal
      $\texttt{messages} \cup \{\texttt{PreVote} \mapsto \texttt{b}\}$ and
      $\texttt{messages} \cup \{\texttt{PreVote} \mapsto \texttt{c}\}$ at once,
      so for any cluster of three or more nodes \texttt{StartPreVote} is
      unsatisfiable and the specification cannot leave its initial state.

\begin{reviewerquote}[reviewer, round 1 --- paraphrased]
Give-up rejected. Your first step is not a transition of \texttt{Next}: two
exact assignments to \texttt{messages'} cannot both hold, so the counterexample
is not an execution of this specification.
\end{reviewerquote}

\noindent
After the review the prover proved Agreement in one further iteration, with the invariant that any node which is Candidate or Leader is the only node in the cluster. The invariant holds for a second reason of the same shape: the only action that makes a node a Candidate conjoins Send, which adds a RequestVote to messages', with Discard, which removes the handled response from it. Whenever another node exists, the two cannot agree. In any cluster of two or more nodes no node ever becomes a Candidate, and Agreement is vacuous.

The reviewer rejected a disproof that was plausible, and steered the prover to
prove both the theorem and the vacuity of the protocol as written. Claiming
that a true property is unprovable is a failure mode we could catch in our pipeline only with a
second, sceptical agent. The two defects it surfaced are both repairs to the
specification rather than to the pipeline: an exact assignment to
\texttt{messages'} composed under a quantifier, a pattern that recurs in our
corpus, and a vote reset at equal term that departs from
Raft.\footnote{The specification does not parse as
published (two operators are used before their definitions) and, once repaired,
violates its own liveness property
\texttt{EventualLeader\,==\,<>(\textbackslash E n \textbackslash in Nodes :
role[n] = "Leader")} by every behaviour in a cluster with three or more nodes.}

% ---------------------------------------------------------------------------

\section{Dataset creation details}
\label{app:dataset_creation}

\subsection{Dataset statistics}

\label{app:dataset-stats}
The two halves of this subsection describe two different populations, and the difference matters for
how each table should be read. \Cref{app:stats-wellposed} covers \emph{every well-posed translation},
i.e.\ the corpus as released. \Cref{app:stats-proven} covers only the \emph{proven} subset: those
translations the prover closed. Nothing in the second half is a statement about the corpus as a whole,
and nothing in the first is conditioned on provability.

\subsubsection{Flattening} The two pipelines produce different artifacts: the algorithmic translator produces one self-contained Verus file per property, while the agentic pipeline produces a Cargo crate with specifications, state machines, and properties across modules. The translations of the agentic pipeline were flattened into a single file, then separated to 3 regions described in \Cref{sec:dataset}.

\subsubsection{All well-posed translations}
\label{app:stats-wellposed}

\begin{wraptable}{r}{0.44\textwidth}
\vspace{-\intextsep}
\centering
\small
\begin{tabular}{lrr}
\toprule
Property type & Algorithmic & Agentic \\
\midrule
safety            &    50 &    630 \\
\texttt{TypeOK}   &   858 & 2{,}286 \\
invariance        & 3{,}440 & 7{,}993 \\
liveness          &   809 & 1{,}443 \\
\midrule
Total             & 5{,}160 & 12{,}352 \\
\bottomrule
\end{tabular}
\caption{\textbf{All well-posed translations}, by property type: every translation that passed the
well-posedness filter, proven or not. The proven subset is counted separately in
\Cref{tab:task-length-proven}.}
\label{tab:by-type}
\vspace{-3em}
\end{wraptable}

\paragraph{Composition.}
\Cref{tab:by-type} gives the well-posed task counts --- every translation that passed the
well-posedness filter, whether or not a proof was later found for it. The agentic encoding yields the
larger corpus in every property type, 12{,}352 tasks against 5{,}160. The two are otherwise balanced
similarly --- invariance is about two thirds of each, \texttt{TypeOK} a sixth --- with one exception:
safety is 1\% of the algorithmic corpus and 5\% of the agentic one.

\subsubsection{The proven subset}
\label{app:stats-proven}

Everything below is restricted to translations the prover closed. On the algorithmic side that is
3{,}094 of the 5{,}160 well-posed tasks in \Cref{tab:by-type}. The agentic side is not the
corresponding subset of its own column: it is a 500-task sample drawn from pairs whose
\emph{algorithmic} translation was already proven, of which 492 were themselves proven. The two $n$
columns of \Cref{tab:task-length-proven} therefore support comparisons down a column but not across.

\paragraph{The eight unclosed pairs.}
Of the eight agentic pairs the prover did not close, four exhausted their budget and four failed due to a
provider-side error. These four were re-run
after the sweep with an identical configuration and all proved in one
iteration; they are excluded from every count and figure in this paper, which
describe the original sweep only.

\paragraph{Task length by property type.}
Every task opens with the same frozen scaffold, so 30{,}780 characters of any length figure is a
shared constant that compresses the difference between the two encodings. \Cref{tab:task-length-proven}
therefore reports both the whole file, which is the length a model actually receives, and the problem
text with the scaffold subtracted. Read on the problem text, three things follow.

First, the agentic encoding is the longer one overall, by 22\%, which is consistent with a translator
that elaborates rather than one working from a fixed set of templates. Second, the gap is not constant
across property types: it is widest on safety and narrows monotonically to liveness, where the two
encodings are the same length to within 1\% and the algorithmic encoding is in fact the marginally
longer of the two. Third, length and difficulty move together here without being the same thing.
Liveness gives the longest problems under both encodings and is the type models solve least often;
safety gives the shortest and is the type they solve most often; yet the algorithmic encoding, which
is the shorter of the two on every type but liveness, is not the easier one.

The safety rows are thin --- 24 proven tasks algorithmic, 2 agentic --- so the 1.51$\times$ at the top
of the ratio column carries the least weight of any row.

\begin{table}[h]
\centering
\small
\begin{tabular}{lrrrrrrr}
\toprule
& \multicolumn{3}{c}{Algorithmic} & \multicolumn{3}{c}{Agentic} & \\
\cmidrule(lr){2-4}\cmidrule(lr){5-7}
Property type & $n$ & whole file & problem & $n$ & whole file & problem & Ratio \\
\midrule
safety            &    24 & 47{,}475 & 16{,}695 &   2 & 56{,}062 & 25{,}282 & 1.51$\times$ \\
\texttt{TypeOK}   &   521 & 47{,}627 & 16{,}847 &  77 & 52{,}509 & 21{,}729 & 1.29$\times$ \\
invariance        & 2{,}245 & 52{,}788 & 22{,}008 & 367 & 57{,}941 & 27{,}161 & 1.23$\times$ \\
liveness          &   304 & 54{,}296 & 23{,}516 &  46 & 53{,}977 & 23{,}197 & 0.99$\times$ \\
\midrule
All               & 3{,}094 & 52{,}026 & 21{,}246 & 492 & 56{,}712 & 25{,}932 & 1.22$\times$ \\
\bottomrule
\end{tabular}
\vspace{1em}
\caption{\textbf{Proven translations only}: counts and mean length, in characters, by property type.
Not a subset relationship with \Cref{tab:by-type} on the agentic side --- see
\Cref{app:stats-proven}. ``Whole file'' is what a model receives; ``problem'' subtracts the frozen
scaffold that opens every task (region~1: 755 lines, 30{,}780 characters, identical corpus-wide).
``Ratio'' is agentic problem length over algorithmic. Dividing by 2.14 gives an approximate token
count; the ratio was measured over the whole corpus and is tokenizer-dependent, so any such figure is
an estimate.}
\label{tab:task-length-proven}
\end{table}

\subsubsection{Source specifications and repositories}
\label{app:corpus-sources}
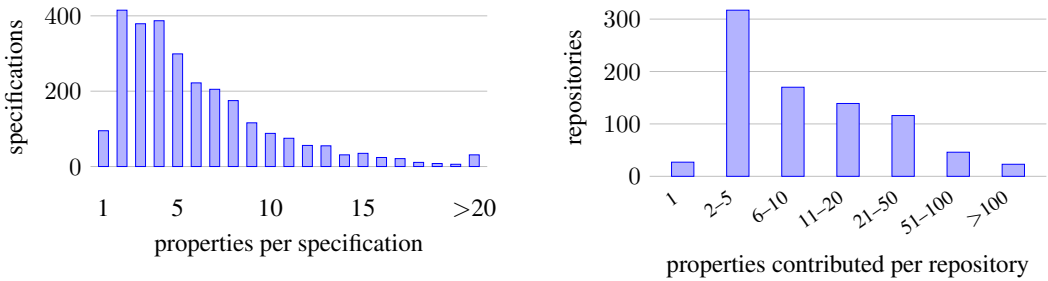
\begin{figure}[h]
  \centering
  \resizebox{\textwidth}{!}{% GENERATED by scripts/paper-corpus-distribution.py — do not edit by hand.
% Requires \usepackage{pgfplots} \pgfplotsset{compat=1.18} in the preamble.
% (a) properties per specification; (b) properties contributed per repository, log-binned. No numbers on the figure: they are in the text.
\begin{tikzpicture}
\begin{axis}[name=a, width=7.2cm, height=4.2cm, ybar, bar width=4pt, xmin=0.3, xmax=21.7,
  xlabel={properties per specification}, ylabel={specifications}, xtick={1,5,10,15,21}, xticklabels={1,5,10,15,{$>$20}},
  ymajorgrids, axis line style={draw=none}, tick style={draw=none}, every axis plot/.append style={fill=tlbluebg,draw=tlblue}]
\addplot coordinates {(1,95) (2,415) (3,379) (4,387) (5,299) (6,222) (7,205) (8,175) (9,116) (10,88) (11,75) (12,56) (13,55) (14,31) (15,35) (16,24) (17,21) (18,11) (19,8) (20,6) (21,31)};
\end{axis}
\begin{axis}[name=b, at={(a.east)}, anchor=west, xshift=2.3cm, width=7.2cm, height=4.2cm, ybar, bar width=9pt,
  xmin=0.4, xmax=7.6, xtick={1,...,7}, xticklabels={{1},{2--5},{6--10},{11--20},{21--50},{51--100},{$>$100}},
  xlabel={properties contributed per repository}, ylabel={repositories}, x tick label style={rotate=35, anchor=east, font=\footnotesize},
  ymajorgrids, axis line style={draw=none}, tick style={draw=none}, every axis plot/.append style={fill=tlgreenbg,draw=tlgreen}]
\addplot coordinates {(1,27) (2,317) (3,170) (4,139) (5,116) (6,46) (7,23)};
\end{axis}
\end{tikzpicture}}
  \caption{(a) How many properties each specification contributes to the
  corpus. (b) How many properties each repository contributes.}
  \label{fig:corpus-sources}
\end{figure}

The 16,459 (specification, property) pairs come from 2,734 TLA$^+$ modules in
838 GitHub repositories (Figure~\ref{fig:corpus-sources}). A module contributes a
median of 5 properties (mean 6.0, maximum 61). 474 repositories contribute one
module; the largest contributes 7.1\% of all pairs, the ten largest 20.5\%, and
67 repositories account for half.

\subsection{Selecting the 40 evaluation tasks}
\textbf{Criterion~3} is three independent checks. \emph{Assumptions}: the generated \verb|assumptions()|
must match the source \verb|ASSUME| with no invented preconditions, since an added hypothesis
makes the theorem strictly weaker and lets a model pass without proving what the specification
asks. \emph{Predicate and polarity}: the same temporal predicate with the same sign, which caught
a pair proving opposite claims about a deliberately buggy specification. \emph{State space}: the
same enumerated vocabulary, retained as metadata rather than used as a filter, because it compares
the two translations to each other when the correct reference is the TLA$^+$ source, and every
pair it excluded turned out to be one where the algorithmic side had under-modelled.

\textbf{Criterion~2} binds hardest on safety: of 630 safety properties with an agentic translation, 578
have no usable algorithmic bundle, leaving 42, of which 24 are proved. Applying criteria 1--6
strictly leaves six safety tasks, so criterion~5 was relaxed for that arm and the affected tasks
carry a flag. The other three arms are unconstrained and were sampled for specification and
repository diversity. Because two properties of one specification may both be selected, the 40
tasks draw on fewer than 40 distinct specifications, and within the safety arm on nine. Selection
never looks at how either encoding is written, only at whether the algorithmic side was proved and
the two agree, so the encoding differences we go on to measure are not themselves a selection
criterion.

The TypeOK and invariance arms are each 10 specifications in 10 repositories,
the liveness arm 8 in 6, the safety arm 9 in 8. The tasks are not evenly spread:
\texttt{jeong-sik/masc-mcp} contributes six, across all four arms, and
\texttt{TheDarkLightX/ZenoDEX} five, four of them liveness and three from a
single specification. Excluding either repository leaves liveness the lowest
type in Table~\ref{tab:by-type-model} on both encodings and changes the model order
of Table~\ref{tab:model-results} only by one adjacent swap each: without
ZenoDEX, GPT-5.6-terra surpasses Qwen3.8-flash on \texttt{tlaverus\_algo};
without masc-mcp, Grok-4.5 reaches 34/34 on \texttt{tlaverus\_llm} and ties
GPT-5.6-sol-pro, ahead on per-sample rate.

\subsection{Sources and licensing}
\label{app:license}
The 40 evaluation tasks are drawn from 32 TLA$^+$ specifications in 24 public GitHub repositories;
we record the repository, commit and SPDX licence of every source file. All are permissively
licensed: 27 tasks come from MIT sources, 12 from Apache-2.0 and one from BSD-2-Clause. Two
repositories are retained despite a contamination flag, because the \emph{safety} arm cannot be
filled otherwise: \texttt{tlaplus/tlaplus}, the canonical TLA$^+$ distribution, and
\texttt{LUC-AI4FM/FormaLLM}, an LLM-formalisation research repository, both of which may appear in
model training data alongside published proofs of the properties we ask for. Beyond the
specifications, the work builds on three third-party assets. The frozen scaffold that opens every
\texttt{tlaverus} task derives from the Anvil verifier
(\url{https://github.com/anvil-verifier/anvil}, Copyright 2022 VMware, Inc., MIT), and its notice is
preserved verbatim inside every task file. Verification uses Verus at the pinned release
\texttt{0.2026.04.19.6f7d4de} on Rust \texttt{1.95.0}. \texttt{verusbench} is an existing benchmark cited in the main text. The evaluated models are accessed through the OpenRouter gateway under the providers' terms
of use. We redistribute none of these assets: the paper reports evaluation results, and neither the
source specifications nor the translated tasks are released with it, so the permissive licences
impose no obligation beyond the attribution given here.

\section{Experimental details}
\label{app:setup}

\paragraph{Models and configuration.}
We evaluate eleven models spanning proprietary and open-weight families.
\Cref{tab:model-config} gives, for each, the parameter count where the developer discloses one, the
endpoint it was served from, and the full decoding configuration. All models are served through a
single API gateway so that prompting, retry and accounting are identical across vendors, and where an
endpoint is named we pin it for the whole run so that samples of one arm are never split across
providers or precisions.

The configuration is uniform by design with only a few places where a model departs from it. Every
run uses temperature $1.0$, $k=3$ samples per task and $V=10$ verification attempts, and no run sets
\texttt{top\_p}, a seed or a chat-template argument. The two \texttt{tlaverus} datasets always share a
model's settings, so the table needs no column to separate them.

\begin{table*}[t]
\centering
\small
\setlength{\tabcolsep}{5pt}
\begin{tabular}{llllcrrl}
\toprule
& & & Pinned provider & & \multicolumn{2}{c}{\texttt{max\_tokens}} & \\
\cmidrule(lr){6-7}
Model & Developer & Params & (\texttt{tlaverus}) & $T$ & \texttt{verusbench} & \texttt{tlaverus} & Effort \\
\midrule
Claude Opus~5    & Anthropic & undisclosed & ---            & 1.0 & 32{,}768 & 128{,}000 & --- \\
Claude Sonnet~5  & Anthropic & undisclosed & ---            & 1.0 & 32{,}768 & 128{,}000 & --- \\
GPT-5.6-sol-pro  & OpenAI    & undisclosed & ---            & 1.0 & 32{,}768 & 128{,}000 & --- \\
GPT-5.6-terra    & OpenAI    & undisclosed & ---            & 1.0 & 32{,}768 & 128{,}000 & --- \\
Grok-4.5         & xAI       & undisclosed & ---            & 1.0 & 32{,}768 & 128{,}000 & --- \\
Grok-4.3         & xAI       & undisclosed & ---            & 1.0 & 32{,}768 & 128{,}000 & high \\
GLM-5.2          & Z.ai      & 744B/40B    & Baseten, FP8   & 1.0 & 32{,}768 & 128{,}000 & --- \\
DeepSeek-R1-0528 & DeepSeek  & 671B/37B    & DeepInfra, FP4 & 1.0 & 32{,}768 & 32{,}768  & --- \\
Qwen3.8-flash    & Alibaba   & undisclosed & ---            & 1.0 & 98{,}304 & 128{,}000 & --- \\
Qwen3.6-35B-a3b  & Alibaba   & 35B/3B      & ---            & 1.0 & 32{,}768 & 128{,}000 & --- \\
gpt-oss-120b     & OpenAI    & 120B/5.1B   & Nebius, FP4    & 1.0 & 32{,}768 & 49{,}152  & --- \\
\bottomrule
\end{tabular}
\vspace{0.5em}
\caption{\textbf{Per-model configuration.} All models are called through OpenRouter. For the
\texttt{tlaverus} runs we pin a specific upstream provider and precision for some open-weight models,
as listed; everywhere else (all \texttt{verusbench} runs, and other model on the \texttt{tlaverus} dataset) we use OpenRouter's default routing at its default precision.
Parameter counts are total/active. A dash is a provider we do not pin or an effort level we do not
set. Grok-4.3's default reasoning effort is low, so we set it high
to match the default reasoning effort of Grok-4.5. DeepSeek-R1 and gpt-oss-120b run below the
\texttt{max\_token = 128000} ceiling on \texttt{tlaverus} because their pinned providers advertise a
smaller one. We set \texttt{max\_token} to 98,304 for Qwen3.8-flash on \texttt{verusbench}, since we
empirically saw that 32,768 would not be enough for this model.}
\label{tab:model-config}
\end{table*}

\paragraph{Budgets.}
The budgets differ by dataset rather than by model,
since a \texttt{tlaverus} task asks for a whole proof region rather than a short function: the
completion limit is 32{,}768 tokens on \texttt{verusbench} against 128{,}000 on \texttt{tlaverus},
the request timeout 600\,s against 1800\,s, and the verifier timeout 60\,s against 300\,s, the last
because a wrong but plausible proof can send the SMT solver far past the time a correct one takes. A
few models depart where an endpoint forces it, e.g. DeepSeek-R1 and gpt-oss-120b run below the \texttt{max\_token = 128,000} ceiling because
their pinned endpoints advertise a smaller one. ; where a model id names a provider and quantisation we hold
that endpoint for the whole run, so the samples of one arm are never split across backends.

\paragraph{What the configuration guarantees.}
These are the values we \emph{send}. Whether a provider honours them is not observable from the
client: a request that sets a temperature the model ignores returns no indication that it did so, so
we report the configuration rather than claim an effect. The one parameter we can check is reasoning
effort, since the provider returns a chain-of-thought token count. Every model with the parameter unset still reports a non-zero count. The setting
therefore scales how much a model reasons rather than switching the behaviour on.

\paragraph{The verification loop.}
A sample proceeds as: render the prompt, take the model's region~3, splice it onto regions~1 and~2,
run Verus. On success the sample ends. On failure the verifier's diagnostics are appended to the
conversation and the model is asked again, to a maximum of $V=10$ verification attempts. A sample
that exhausts the budget is recorded as \texttt{verif\_cap} and scored a failure. Since each attempt
is one model call, a sample costs up to $V+1$ calls, and mean turns per sample is reported alongside
accuracy as a measure of how quickly a model converges.

\paragraph{Verifier and toolchain.}
All datasets are graded with Verus at a single pinned release, \texttt{0.2026.04.19.6f7d4de}, on Rust
\texttt{1.95.0}, inside a container image built once per dataset. The pin matters: Verus changes the
accepted surface syntax between releases, and a later toolchain fails every task in this corpus at
compile time regardless of the proof. Pinning also makes a score reproducible, since the same
submission can change verdict under a different prover version.

\paragraph{Ground-truth separation.}
Model-facing jobs read a mirror of the dataset from which the reference proofs have been removed, and
run under a credential that is denied access to the prefix holding them. A job that tried to read the
reference proofs would fail with a permission error rather than silently succeed, so a leak cannot
occur unobserved.

\paragraph{Reporting.}
We report pass@$k$ at $k=3$, the per-sample pass rate over all samples drawn, mean model calls per
sample, and mean reasoning and answer tokens per sample. Reasoning tokens are the part of the
provider's completion count reported separately as chain of thought; where an endpoint does not
report them the split is marked unavailable rather than assumed to be zero. Cost is the provider's
own billed figure summed over the run.

\section{Further evaluation results}
\label{app:eval_results}

\begin{figure*}[h]
\centering
\includegraphics[width=\linewidth]{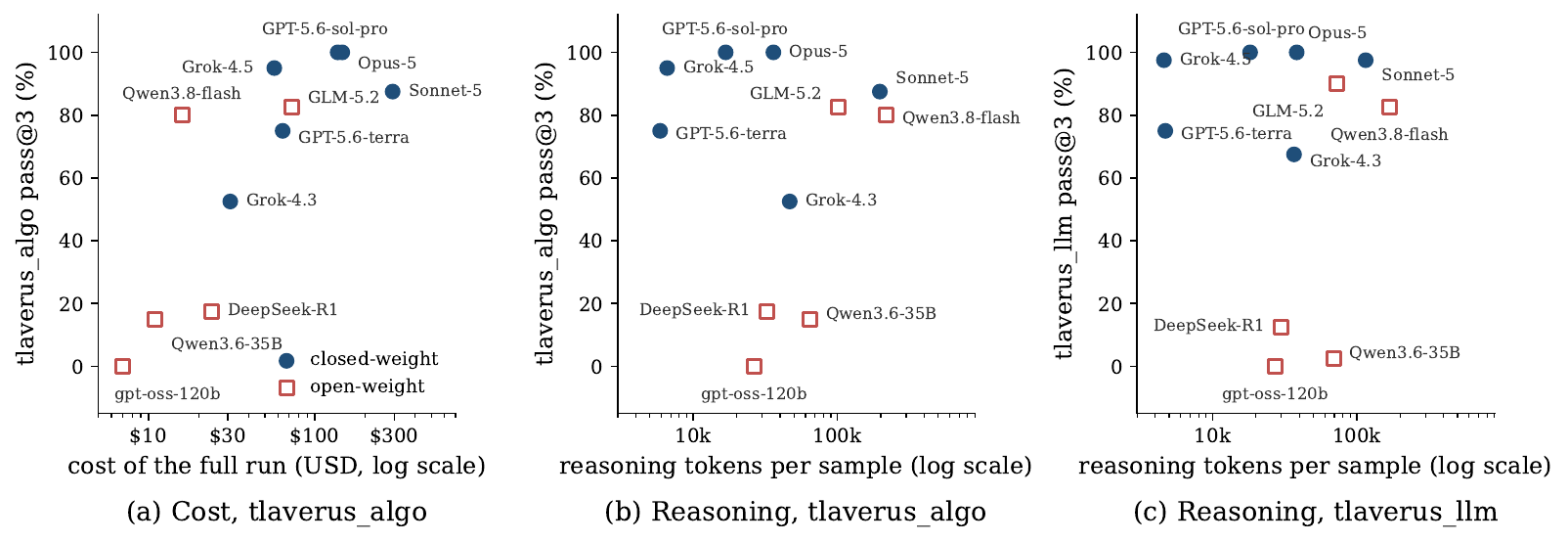}
\caption{\textbf{(a)} Cost of the full \texttt{tlaverus\_algo} run against pass@3; a run is 40 tasks
$\times$ 3 samples. \textbf{(b, c)} Reasoning tokens per sample against pass@3 on the two \texttt{tlaverus} datasets. All axes but pass@3 are logarithmic.}
\label{fig:cost-reasoning}
\end{figure*}

\paragraph{Cost and reasoning in full.}
\Cref{fig:cost-reasoning} gives the two views the main text draws on. Panel~a is the cost plot for
\texttt{tlaverus\_algo}, the dataset omitted from \Cref{fig:results-overview}b; it shows the same
picture, with the same models off the line. Grok-4.5 beats Sonnet-5
outright here, 95.0\% for \$57 against 87.5\% for \$293, a sharper version of the tie it reaches on
\texttt{tlaverus\_llm}. Per solved task the field spans \$0.50 for Qwen3.8-flash and \$1.50 for
Grok-4.5 against \$3.65 for Opus-5 and \$8.37 for Sonnet-5.

Panels~b and~c plot reasoning tokens against pass@3 on the two datasets.
The two are uncorrelated on both (Spearman $\rho = -0.06$ on \texttt{tlaverus\_algo},
$\rho = 0.00$ on \texttt{tlaverus\_llm}; 11 models each), so the finding holds
across two encodings of the same 40 \mbox{(spec, property)} pairs. Sonnet-5 spends 196.9k reasoning
tokens per sample on \texttt{tlaverus\_algo} and scores below Grok-4.5 at 6.6k; Qwen3.8-flash spends
more than any other model, 217.9k, to land mid-field at 80.0\%; gpt-oss-120b spends 26.6k to solve
nothing. Reasoning and cost are uncorrelated with each other as well, so
the models deliberating most are neither the priciest nor the most accurate.

\begin{table}[h]
\centering
\small
\setlength{\tabcolsep}{5pt}
\begin{tabular}{lrrrrrr}
\toprule
Model & p@3 & p@1 & turns & reas. & ans. & cost \\
\midrule
Opus-5          & 99.3 & 99.3 & 1.4 & 3.4k & 1.1k &  \$54 \\
Sonnet-5        & 98.7 & 94.2 & 2.6 & 10.6k & 2.0k &  \$67 \\
GPT-5.6-sol-pro & 97.3 & 93.3 & 3.3 & 10.9k & 7.8k & \$167 \\
Grok-4.5        & 95.3 & 90.4 & 2.6 & 3.7k & 1.9k &  \$17 \\
GLM-5.2         & 88.7 & 75.6 & 5.6 & 23.7k & 5.9k &  \$53 \\
Grok-4.3        & 86.0 & 78.2 & 4.7 & 22.9k & 2.5k &  \$31 \\
GPT-5.6-terra   & 85.3 & 73.1 & 5.6 & 3.8k & 4.5k &  \$55 \\
Qwen3.8-flash   & 80.7 & 58.2 & 6.7 &  353.9k & 9.4k &   \$77 \\
DeepSeek-R1     & 74.7 & 57.6 & 6.3 & 24.3k & 5.7k &  \$31 \\
gpt-oss-120b    & 29.3 & 13.1 & 9.2 & 9.0k & 4.3k &   \$3 \\
Qwen3.6-35B     & 28.0 & 14.4 & 4.4 & 18.9k & 6.3k &  \$23 \\
\bottomrule
\end{tabular}
\vspace{0.5em}
\caption{\textbf{\texttt{verusbench}}, 150 tasks. Rows ordered by p@3.}
\label{tab:full-vb}
\end{table}

\begin{table}[h]
\centering
\small
\setlength{\tabcolsep}{5pt}
\begin{tabular}{lrrrrrr}
\toprule
Model & p@3 & p@1 & turns & reas. & ans. & cost \\
\midrule
Opus-5          & 100.0 & 100.0 &  1.1 &  36.0k & 7.8k & \$146 \\
GPT-5.6-sol-pro & 100.0 &  92.5 &  3.5 &  16.8k & 46.1k & \$137 \\
Grok-4.5        &  95.0 &  90.8 &  4.2 &   6.6k & 44.7k &  \$57 \\
Sonnet-5        &  87.5 &  82.5 &  3.4 & 196.9k & 18.8k & \$293 \\
GLM-5.2         &  82.5 &  70.8 &  5.7 & 101.6k & 26.0k &  \$73 \\
Qwen3.8-flash   &  80.0 &  68.3 &  6.4 & 217.9k & 49.8k &  \$16 \\
GPT-5.6-terra   &  75.0 &  60.8 &  6.8 &   5.9k & 15.5k &  \$64 \\
Grok-4.3        &  52.5 &  40.8 &  8.0 &  46.7k & 20.1k &  \$31 \\
DeepSeek-R1     &  17.5 &  10.8 &  9.6 &  32.4k & 13.5k &  \$24 \\
Qwen3.6-35B     &  15.0 &   5.0 &  9.9 &  64.3k & 16.9k &  \$11 \\
gpt-oss-120b    &   0.0 &   0.0 & 10.0 &  26.6k &  8.4k &   \$7 \\
\bottomrule
\end{tabular}
\vspace{0.5em}
\caption{\textbf{\texttt{tlaverus\_algo}}, 40 tasks. Rows ordered by p@3.}
\label{tab:full-algo}
\end{table}

\begin{table}[h]
\centering
\small
\setlength{\tabcolsep}{5pt}
\begin{tabular}{lrrrrrr}
\toprule
Model & p@3 & p@1 & turns & reas. & ans. & cost \\
\midrule
Opus-5          & 100.0 & 100.0 &  1.6 &  38.3k & 7.4k & \$171 \\
GPT-5.6-sol-pro & 100.0 &  83.3 &  5.5 &  18.2k & 51.5k & \$180 \\
Grok-4.5        &  97.5 &  90.0 &  5.0 &   4.6k & 36.8k &  \$51 \\
Sonnet-5        &  97.5 &  87.5 &  4.6 & 115.2k & 19.4k & \$214 \\
GLM-5.2         &  90.0 &  78.3 &  5.6 &  72.5k & 18.9k &  \$55 \\
Qwen3.8-flash   &  82.5 &  60.0 &  7.3 & 169.6k & 40.8k &  \$14 \\
GPT-5.6-terra   &  75.0 &  52.5 &  7.5 &   4.7k & 16.4k &  \$70 \\
Grok-4.3        &  67.5 &  52.5 &  7.6 &  36.7k & 12.6k &  \$27 \\
DeepSeek-R1     &  12.5 &   7.5 &  9.8 &  29.8k & 15.6k &  \$26 \\
Qwen3.6-35B     &   2.5 &   0.8 & 10.0 &  69.0k & 20.1k &  \$13 \\
gpt-oss-120b    &   0.0 &   0.0 & 10.0 &  27.2k & 10.2k &   \$8 \\
\bottomrule
\end{tabular}
\vspace{0.5em}
\caption{\textbf{\texttt{tlaverus\_llm}}, 40 tasks. Rows ordered by p@3. \emph{p@3} is the share of
tasks solved by at least one of $k=3$ samples, \emph{p@1} the share of all samples that passed,
\emph{turns} the mean model calls per sample, \emph{reas.} and \emph{ans.} the mean disjoint
reasoning and answer tokens per sample, and \emph{cost} the price of the whole run. The same columns
appear in \Cref{tab:full-vb,tab:full-algo}.}
\label{tab:full-llm}
\end{table}

\paragraph{Full per-model results.}
\Cref{tab:full-vb,tab:full-algo,tab:full-llm} give every measured quantity for the three datasets,
one table each. The accuracy columns are those of \Cref{tab:model-results}; what follows concerns the
rest.

\emph{Answer length is the quantity that changes most between benchmarks.} A \texttt{verusbench}
answer averages 1.1k to 9.4k tokens per sample; a \texttt{tlaverus} answer 7.4k to 51.5k. The
increase is not uniform, and it is largest for the models that do best: on \texttt{tlaverus\_algo} Grok-4.5 grows $24\times$ (1.9k to 44.7k) and Sonnet-5 $9\times$, while gpt-oss-120b grows $2\times$ and DeepSeek-R1 $2.4\times$; Opus-5 ($7\times$, 1.1k to 7.8k) is the exception, writing the shortest \texttt{tlaverus} answers of any model. That is the proof body getting longer rather than the search for it getting harder ---
a weak model cannot write a long proof it has not found, so its answers stay short no matter how
many attempts it makes.

\emph{The split between reasoning and answer is a property of the model, not of the task.} The share
of the completion reported as reasoning is nearly identical on the two \texttt{tlaverus} datasets for
every model --- Sonnet-5 at 91\% and 86\%, Qwen3.8-flash at 81\% and 81\%, Grok-4.5 at 13\% and
11\%, GPT-5.6-terra at 28\% and 22\%. A model that deliberates before writing does so on both
encodings, in about the same proportion, which is why reasoning volume carries no information about
either dataset.

\emph{Turns rise almost everywhere, and cost rises for all but one model once normalised.} Every model needs
more model calls per sample on \texttt{tlaverus} than on \texttt{verusbench} on at least one dataset; on \texttt{tlaverus\_algo} Opus-5 (1.1 against 1.4) and Qwen3.8-flash (6.4 against 6.7) need marginally fewer, and GLM-5.2 stays at 5.6--5.7 throughout. The \emph{cost} column is not comparable across datasets as
printed, since a \texttt{verusbench} run covers 450 samples and a \texttt{tlaverus} run 120; per sample the price rises for every model except Qwen3.8-flash, whose \texttt{tlaverus} runs cost $0.8\times$ and $0.7\times$ its \texttt{verusbench} rate, from $1.8\times$ for Qwen3.6-35B to $16\times$ for Sonnet-5 on \texttt{tlaverus\_algo}.
That factor tracks the $77$--$89\times$ difference in prompt length only loosely, because the prompt
is re-sent on each turn and the turn counts differ.

\begin{table}[h]
\centering
\small
\setlength{\tabcolsep}{5pt}
\begin{tabular}{lrrrrrrrr}
\toprule
& \multicolumn{4}{c}{\texttt{tlaverus\_algo}} & \multicolumn{4}{c}{\texttt{tlaverus\_llm}} \\
\cmidrule(lr){2-5}\cmidrule(lr){6-9}
Model & safe & \texttt{TypeOK} & inv & live & safe & \texttt{TypeOK} & inv & live \\
\midrule
Opus-5          & 10 & 10 & 10 & 10 & 10 & 10 & 10 & 10 \\
GPT-5.6-sol-pro & 10 & 10 & 10 & 10 & 10 & 10 & 10 & 10 \\
Grok-4.5        & 10 &  8 & 10 & 10 & 10 & 10 & 10 &  9 \\
Sonnet-5        & 10 &  8 & 10 &  7 & 10 &  9 & 10 & 10 \\
GLM-5.2         & 10 &  8 &  9 &  6 & 10 & 10 &  9 &  7 \\
Qwen3.8-flash   &  8 &  8 &  8 &  8 &  9 &  7 &  9 &  8 \\
GPT-5.6-terra   &  8 &  8 & 10 &  4 &  8 &  8 &  9 &  5 \\
Grok-4.3        &  7 &  8 &  4 &  2 &  9 &  9 &  7 &  2 \\
DeepSeek-R1     &  0 &  4 &  3 &  0 &  1 &  2 &  1 &  1 \\
Qwen3.6-35B     &  3 &  1 &  1 &  1 &  1 &  0 &  0 &  0 \\
gpt-oss-120b    &  0 &  0 &  0 &  0 &  0 &  0 &  0 &  0 \\
\midrule
All             & 76 & 73 & 75 & 58 & 78 & 75 & 75 & 62 \\
\bottomrule
\end{tabular}
\vspace{1em}
\caption{\textbf{Tasks solved by property type}, out of 10 per cell; the bottom row is out of 110
model--task pairs. The two datasets hold the same 40 pairs and are comparable cell by cell.}
\label{tab:by-type-model}
\end{table}

\paragraph{Property type per model.}
\Cref{tab:by-type-model} gives the cells behind the aggregate in \Cref{fig:results-overview}c. Two
patterns survive at the level of individual models. First, liveness is the type that separates the
field: it is strictly the worst type for four models on each dataset, and the only other strict minima are on \texttt{TypeOK}, by one or two tasks (Grok-4.5 on \texttt{tlaverus\_algo}; Sonnet-5 and Qwen3.8-flash on \texttt{tlaverus\_llm}). The break is not confined to weak models --- Sonnet-5 holds 10/10 on safety and
invariance on \texttt{tlaverus\_algo} and drops to 7/10 on liveness, and GPT-5.6-terra goes 10/10 to
4/10.
 
Second, liveness is where the two encodings diverge least. Grok-4.3 gains on three types under the
agentic translation, safety 7 to 9, \texttt{TypeOK} 8 to 9 and invariance 4 to 7, and stays at 2/10
on liveness; GPT-5.6-terra moves by one task there and by at most one elsewhere. Sonnet-5 is the
exception, going 7 to 10, which accounts for most of its four-task gain. Whatever makes liveness hard
is therefore largely a property of what is being proved rather than of how the specification was
translated, which is consistent with the two datasets differing by under 2\% in problem-text length on
this type (\Cref{tab:dataset-size-bytype}).

\section{Failure analysis of unresolved tasks}
\label{app:failure-analysis}

We call a task \emph{unresolved} for a model when none of its three samples passes. Across the runs
behind Tables~11--13 this gives 355 unresolved model--task pairs on verusbench, 158 on
tlaverus\_algo and 150 on tlaverus\_llm (1{,}989 failed samples). We classify every failed sample
once, by its final attempt, using the verifier output, the reply's finish reason and the anti-cheat
verdict recorded in the trace (first matching category wins: output-token limit, solver budget,
specification edit, no answer, refuted proof obligation, trigger, typing/mode, unknown names, other
syntax). Table~\ref{tab:fail-pooled} pools the models; Tables~\ref{tab:fail-verusbench},
\ref{tab:fail-tlaverus-algo} and~\ref{tab:fail-tlaverus-llm} give the per-model breakdown.

\paragraph{Proof budget.} Budget limits rarely end a sample, but the attempt budget is almost always
spent. Only 1--2\% of final attempts were cut at the output-token limit, concentrated in a few
models (Sonnet-5 on the tlaverus sets, GLM-5.2 on verusbench), and no unresolved sample ended on a
solver timeout or resource limit. In contrast, 95\% (verusbench) and 99\% (both tlaverus sets) of
the failed samples used all 10 verifier checks. Unresolved tasks are therefore not cut short by
compute: the models keep producing proofs that do not check until the attempt budget runs out.
The output-token limit was 128k on tlaverus (32{,}768 for DeepSeek-R1 and 49{,}152 for
gpt-oss-120b) and 32{,}768 on verusbench (64{,}000 for Opus-5 and 98{,}304 for Qwen3.8-flash).

\paragraph{Triggers.} Trigger errors (Verus cannot infer, or rejects, a quantifier trigger) are rare
on verusbench (2\%) but are 13--15\% of failures on the tlaverus sets, and 24\% on liveness
properties. They are a major cause for mid-ranked models on tlaverus (37--43\% for GPT-5.6-terra, 33--67\% for
GLM-5.2). The tlaverus proofs quantify over executions and states, which
the verusbench programs rarely require.

\paragraph{Typing.} Type and mode errors show the opposite pattern: 23\% of verusbench failures but
only 4--6\% on tlaverus. On verusbench about half are mismatched types and a fifth are mathematical types
(\texttt{int}, \texttt{nat}) used in executable code; the tlaverus tasks ask for proof functions and
contain little executable code.

\paragraph{Missing environment assumptions.} We find no evidence that tasks fail because an
environment assumption is missing from the task statement. Every tlaverus task is solved, under the same anti-cheat check, by at least
two models, so the hypotheses given in the task suffice for a proof. Attempts that verify only after editing the task's code or specification, which is how a
model would add an assumption it believes is missing, are 0--1\% of tlaverus failures (5\% on verusbench). What the data does show is that
\emph{using} the given assumptions is hard: liveness properties, whose proofs must reason over whole
executions using the fairness hypotheses, are unresolved in 45\% of model--task
pairs, against 30--33\% for invariance, safety and type-correctness properties, and account for most
trigger failures.

\paragraph{Other causes.} The largest category is other syntax errors (44--48\%), driven by the
weaker models (up to 69\% for gpt-oss-120b on verusbench and 82\% on both tlaverus sets); on tlaverus a further 10--13\% call
lemmas, functions or types that do not exist. Refuted proof obligations, the failures closest to a
genuine reasoning gap, are 17\% of verusbench and 21--24\% of tlaverus failures, and the leading
category for the strongest models' few unresolved tlaverus tasks (Grok-4.5).

\begin{table}[t]
\centering
\small
\setlength{\tabcolsep}{3.5pt}
\begin{tabular}{lrrrrrrrrrr}
\toprule
 & & & \multicolumn{7}{c}{final attempt (\% of samples)} & \\
\cmidrule(lr){4-10}
Benchmark & unres. & samples & token & spec & trigger & typing & names & syntax & proof & cap \\
\midrule
verusbench & 355 & 1065 & 2 & 5 & 2 & 23 & 4 & 48 & 17 & 95 \\
tlaverus\_algo & 158 & 474 & 1 & 1 & 13 & 4 & 13 & 44 & 24 & 99 \\
tlaverus\_llm & 150 & 450 & 1 & 0 & 15 & 6 & 10 & 47 & 21 & 99 \\
\midrule
\quad invariance & 70 & 210 & 0 & 0 & 10 & 3 & 10 & 50 & 26 & 99 \\
\quad safety\_typeOK & 72 & 216 & 3 & 0 & 6 & 5 & 11 & 51 & 23 & 99 \\
\quad safety & 66 & 198 & 0 & 1 & 11 & 5 & 16 & 48 & 20 & 99 \\
\quad liveness & 100 & 300 & 1 & 1 & 24 & 6 & 11 & 36 & 22 & 99 \\
\bottomrule
\end{tabular}
\vspace{1em}
\caption{Final-attempt failure categories of unresolved tasks (a model--task pair with 0 of 3 samples passing), pooled over models. Each failed sample is classified once, by its last attempt. \emph{token}: the reply was cut at the output-token limit; \emph{spec}: the proof verified but the anti-cheat rejected an edit to the task's code or specification; \emph{trigger}: Verus could not infer, or rejected, a quantifier trigger; \emph{typing}: type and mode errors (e.g.\ \texttt{int}/\texttt{nat} in executable code, mismatched types, spec/proof/exec mode); \emph{names}: calls to functions, types or lemmas that do not exist; \emph{syntax}: other parse and Verus-syntax errors; \emph{proof}: Verus checked the proof and refuted an obligation. \emph{cap}: samples that used all 10 verifier checks. Solver timeouts (rlimit), empty replies and verifier crashes account for 3 of 1989 samples and are not shown. The lower block splits both tlaverus sets by property type.}
\label{tab:fail-pooled}
\end{table}

\begin{table}[t]
\centering
\small
\setlength{\tabcolsep}{3.5pt}
\begin{tabular}{lrrrrrrrrrr}
\toprule
 & & & \multicolumn{7}{c}{final attempt (\% of samples)} & \\
\cmidrule(lr){4-10}
Model & unres. & samples & token & spec & trigger & typing & names & syntax & proof & cap \\
\midrule
Opus-5 & 1 & 3 & 0 & 0 & 0 & 100 & 0 & 0 & 0 & 100 \\
GPT-5.6-sol-pro & 4 & 12 & 0 & 0 & 8 & 17 & 8 & 42 & 25 & 100 \\
Sonnet-5 & 2 & 6 & 0 & 0 & 0 & 50 & 50 & 0 & 0 & 100 \\
Grok-4.5 & 7 & 21 & 0 & 24 & 10 & 19 & 0 & 5 & 43 & 76 \\
GLM-5.2 & 17 & 51 & 10 & 14 & 0 & 27 & 2 & 12 & 35 & 86 \\
Qwen3.8-flash & 29 & 87 & 2 & 6 & 2 & 26 & 6 & 45 & 13 & 94 \\
GPT-5.6-terra & 22 & 66 & 0 & 27 & 5 & 17 & 0 & 18 & 33 & 73 \\
Grok-4.3 & 21 & 63 & 0 & 5 & 6 & 24 & 0 & 6 & 59 & 95 \\
DeepSeek-R1 & 38 & 114 & 0 & 10 & 7 & 26 & 5 & 25 & 27 & 91 \\
Qwen3.6-35B & 108 & 324 & 3 & 1 & 0 & 24 & 5 & 59 & 8 & 99 \\
gpt-oss-120b & 106 & 318 & 0 & 2 & 0 & 19 & 2 & 69 & 7 & 98 \\
\bottomrule
\end{tabular}
\vspace{1em}
\caption{Final-attempt failure categories of unresolved tasks per model, verusbench, 150 tasks. Columns as in Table~\ref{tab:fail-pooled}; models with no unresolved task are omitted.}
\label{tab:fail-verusbench}
\end{table}
\vspace{-1em}

\begin{table}[t]
\centering
\small
\setlength{\tabcolsep}{3.5pt}
\begin{tabular}{lrrrrrrrrrr}
\toprule
 & & & \multicolumn{7}{c}{final attempt (\% of samples)} & \\
\cmidrule(lr){4-10}
Model & unres. & samples & token & spec & trigger & typing & names & syntax & proof & cap \\
\midrule
Sonnet-5 & 5 & 15 & 40 & 0 & 40 & 0 & 7 & 0 & 13 & 100 \\
Grok-4.5 & 2 & 6 & 0 & 0 & 0 & 0 & 0 & 0 & 100 & 100 \\
GLM-5.2 & 7 & 21 & 0 & 0 & 33 & 0 & 0 & 24 & 43 & 100 \\
Qwen3.8-flash & 8 & 24 & 0 & 4 & 38 & 8 & 4 & 29 & 17 & 96 \\
GPT-5.6-terra & 10 & 30 & 0 & 0 & 37 & 3 & 3 & 13 & 43 & 100 \\
Grok-4.3 & 19 & 57 & 0 & 0 & 11 & 4 & 18 & 21 & 46 & 98 \\
DeepSeek-R1 & 33 & 99 & 0 & 1 & 10 & 1 & 24 & 43 & 20 & 99 \\
Qwen3.6-35B & 34 & 102 & 0 & 0 & 9 & 12 & 17 & 38 & 25 & 100 \\
gpt-oss-120b & 40 & 120 & 0 & 1 & 2 & 1 & 7 & 82 & 8 & 99 \\
\bottomrule
\end{tabular}
\vspace{1em}
\caption{Final-attempt failure categories of unresolved tasks per model, tlaverus\_algo, 40 tasks. Columns as in Table~\ref{tab:fail-pooled}; models with no unresolved task are omitted.}
\label{tab:fail-tlaverus-algo}
\end{table}

\begin{table}[t]
\centering
\small
\setlength{\tabcolsep}{3.5pt}
\begin{tabular}{lrrrrrrrrrr}
\toprule
 & & & \multicolumn{7}{c}{final attempt (\% of samples)} & \\
\cmidrule(lr){4-10}
Model & unres. & samples & token & spec & trigger & typing & names & syntax & proof & cap \\
\midrule
Sonnet-5 & 1 & 3 & 100 & 0 & 0 & 0 & 0 & 0 & 0 & 100 \\
Grok-4.5 & 1 & 3 & 0 & 0 & 0 & 0 & 0 & 0 & 100 & 100 \\
GLM-5.2 & 4 & 12 & 0 & 0 & 67 & 0 & 8 & 17 & 8 & 100 \\
Qwen3.8-flash & 7 & 21 & 0 & 0 & 29 & 0 & 14 & 19 & 38 & 100 \\
GPT-5.6-terra & 10 & 30 & 0 & 0 & 43 & 10 & 0 & 10 & 37 & 100 \\
Grok-4.3 & 13 & 39 & 0 & 0 & 33 & 3 & 5 & 18 & 41 & 100 \\
DeepSeek-R1 & 35 & 105 & 0 & 1 & 10 & 8 & 23 & 35 & 21 & 97 \\
Qwen3.6-35B & 39 & 117 & 0 & 1 & 13 & 13 & 7 & 50 & 16 & 99 \\
gpt-oss-120b & 40 & 120 & 0 & 0 & 0 & 0 & 6 & 82 & 12 & 100 \\
\bottomrule
\end{tabular}
\vspace{1em}
\caption{Final-attempt failure categories of unresolved tasks per model, tlaverus\_llm, 40 tasks. Columns as in Table~\ref{tab:fail-pooled}; models with no unresolved task are omitted.}
\label{tab:fail-tlaverus-llm}
\end{table}

\section{Compute resources}
\label{app:compute} 
Dataset construction consumed approximately \$16{,}200 of Claude tokens for agentic translation and \$13{,}300 for proof search across both pipelines, so approximately \$29{,}500 total, valued at list API prices; the work was run under a Claude subscription, so this is a measure of compute used rather than an amount billed. The algorithmic transpiler itself uses no model calls. Evaluation costs are reported per model in \Cref{app:eval_results}, making it approximately \$2{,}500 in total.

%% GENERATED from fakellm-analysis-2026-09-09 (balazs/appendix.tex). Figures: balazs/figures/*.tex, data: balazs/data/.
%% Needs \input{balazs/figures/common} in the main preamble.
%% Fallbacks in case the main preamble does not define these (no-ops if it does).
\providecommand{\code}[1]{\texttt{#1}}
\providecommand{\draftmark}{}
\section{LLM versus algorithmic translations}
\label{app:llm-alg}

% Provenance (not for publication):
%   tlaverus_proof_tracker rows sk=*#fakellm (500 LLM bundles staged as
%   outputs/transpec-fakellm000.../, 2026-09-08) joined to their sk=*#8ed60c2 twin
%   on (tla_file_sha, property). Both sides proven by the unchanged Claude bundle
%   pipeline (Opus 4.8 prover, Sonnet 5 reviewer). 492/500 LLM proven
%   (4 iterations_exhausted, 4 rate_limited). Solver numbers: local re-verification
%   of each final bundle.rs with Verus 0.2026.04.19 (--time-expanded --output-json,
%   --profile-all). Snapshot 2026-09-09. Raw table: balazs/data/pairs.csv (not in repo).

\begin{figure}[H]
\centering
\begin{subfigure}[b]{0.51\linewidth}
\centering
%% Paired ratios LLM / algorithmic: median of per-pair log2 ratios with 95% bootstrap interval.
%% Values inlined from balazs/data/fig2_forest.csv. Restyled 2026-09-14 to match Fig. 4 (navy, gray axes);
%% group headings are extra y ticks. Fragment: \input from the main document.
\begin{tikzpicture}
\begin{axis}[base, width=5.5cm, height=7.2cm, xmode=log, log basis x=2,
  xmin=0.125, xmax=8, ymin=0.4, ymax=15.6,
  xtick={0.125,0.25,0.5,1,2,4,8},
  xticklabels={$\nicefrac18\times$,$\nicefrac14\times$,$\nicefrac12\times$,$1\times$,$2\times$,$4\times$,$8\times$},
  ytick={15,14,13,12,11,10,9,8,7,6,5,4,3,2,1},
  yticklabels={{\color{black!55}\scshape prover effort},{proof cost},{prover turns},{verus runs},{context read},
               {\color{black!55}\scshape translation},{translation lines},{comment lines},{$\forall$ quantifiers},{longest line},
               {\color{black!55}\scshape proof},{proof lines},
               {\color{black!55}\scshape solver},{Z3 rlimit},{Z3 instantiations}},
  y tick label style={font=\footnotesize, align=right},
  xlabel={LLM $\div$ algorithmic}, xlabel style={font=\footnotesize},
  grid=none, clip=false, axis y line*=left, y axis line style={draw=none}, axis x line*=bottom,
  ytick style={draw=none}, ytick align=inside]
\draw[gridc] (axis cs:0.25,0.4)--(axis cs:0.25,15.6);
\draw[gridc] (axis cs:0.5,0.4)--(axis cs:0.5,15.6);
\draw[gridc] (axis cs:2,0.4)--(axis cs:2,15.6);
\draw[gridc] (axis cs:4,0.4)--(axis cs:4,15.6);
\draw[black!55] (axis cs:1,0.4)--(axis cs:1,15.6);
\draw[alg, line width=1.2pt] (axis cs:1.1538,14)--(axis cs:1.2303,14); \fill[alg] (axis cs:1.1901,14) circle (1.9pt);
\node[anchor=west, font=\footnotesize, text=ink] at (axis cs:8,14) {1.19$\times$};
\draw[alg, line width=1.2pt] (axis cs:1.3,13)--(axis cs:1.4444,13); \fill[alg] (axis cs:1.375,13) circle (1.9pt);
\node[anchor=west, font=\footnotesize, text=ink] at (axis cs:8,13) {1.38$\times$};
\draw[alg, line width=1.2pt] (axis cs:1.5,12)--(axis cs:2.0,12); \fill[alg] (axis cs:2.0,12) circle (1.9pt);
\node[anchor=west, font=\footnotesize, text=ink] at (axis cs:8,12) {2.00$\times$};
\draw[alg, line width=1.2pt] (axis cs:1.737,11)--(axis cs:1.9958,11); \fill[alg] (axis cs:1.882,11) circle (1.9pt);
\node[anchor=west, font=\footnotesize, text=ink] at (axis cs:8,11) {1.88$\times$};
\draw[alg, line width=1.2pt] (axis cs:2.3739,9)--(axis cs:2.4198,9); \fill[alg] (axis cs:2.3899,9) circle (1.9pt);
\node[anchor=west, font=\footnotesize, text=ink] at (axis cs:8,9) {2.39$\times$};
\draw[alg, line width=1.2pt] (axis cs:5.3125,8)--(axis cs:5.4468,8); \fill[alg] (axis cs:5.3652,8) circle (1.9pt);
\node[anchor=west, font=\footnotesize, text=ink] at (axis cs:8,8) {5.37$\times$};
\draw[alg, line width=1.2pt] (axis cs:0.3061,7)--(axis cs:0.375,7); \fill[alg] (axis cs:0.3333,7) circle (1.9pt);
\node[anchor=west, font=\footnotesize, text=ink] at (axis cs:8,7) {0.33$\times$};
\draw[alg, line width=1.2pt] (axis cs:0.1347,6)--(axis cs:0.1625,6); \fill[alg] (axis cs:0.1487,6) circle (1.9pt);
\node[anchor=west, font=\footnotesize, text=ink] at (axis cs:8,6) {0.15$\times$};
\draw[alg, line width=1.2pt] (axis cs:1.01,4)--(axis cs:1.075,4); \fill[alg] (axis cs:1.0482,4) circle (1.9pt);
\node[anchor=west, font=\footnotesize, text=ink] at (axis cs:8,4) {1.05$\times$};
\draw[alg, line width=1.2pt] (axis cs:0.8254,2)--(axis cs:0.915,2); \fill[alg] (axis cs:0.86,2) circle (1.9pt);
\node[anchor=west, font=\footnotesize, text=ink] at (axis cs:8,2) {0.86$\times$};
\draw[alg, line width=1.2pt] (axis cs:0.6111,1)--(axis cs:0.7875,1); \fill[alg] (axis cs:0.7273,1) circle (1.9pt);
\node[anchor=west, font=\footnotesize, text=ink] at (axis cs:8,1) {0.73$\times$};
\end{axis}
\end{tikzpicture}
\caption{Paired ratios}
\label{fig:llm-alg-ratios}
\end{subfigure}\hfill
\begin{subfigure}[b]{0.47\linewidth}
\centering
%% What each verus run came back with: mean runs per proof by outcome class.
%% Reads balazs/data/fig10_run_outcomes.csv. Restyled 2026-09-14 to match Fig. 4. Fragment: \input from the main document.
\begin{tikzpicture}
\begin{axis}[base, width=5.2cm, height=7.2cm, xbar, bar width=6pt,
  symbolic y coords={no_result_shown,other_compile_error,name_resolution_error,trigger_error,verification_failed,verified},
  ytick={no_result_shown,other_compile_error,name_resolution_error,trigger_error,verification_failed,verified},
  yticklabels={no result,other compile error,name resolution error,trigger error,verification failed,verified},
  y tick label style={font=\footnotesize},
  xmin=0, xmax=2.05, xtick={0,0.5,1,1.5,2}, xlabel={verus runs per proof (mean)}, xlabel style={font=\footnotesize},
  xmajorgrids=true, grid style={gridc, thin}, axis y line*=left, axis x line*=bottom,
  nodes near coords={\pgfmathprintnumber[fixed,precision=2]{\pgfplotspointmeta}},
  nodes near coords style={font=\scriptsize, text=ink},
  legend style={font=\footnotesize, at={(0.97,0.22)}, anchor=east}, enlarge y limits=0.1]
\addplot[fill=llm, draw=none] table[x=llm_mean_runs, y=outcome, col sep=comma] {balazs/data/fig10_run_outcomes.csv};
\addlegendentry{LLM}
\addplot[fill=alg, draw=none] table[x=alg_mean_runs, y=outcome, col sep=comma] {balazs/data/fig10_run_outcomes.csv};
\addlegendentry{algorithmic}
\pgfplotsset{reverse legend}
\end{axis}
\end{tikzpicture}
\caption{Outcome of each verus run}
\label{fig:llm-alg-verus-runs}
\end{subfigure}
\caption{\textbf{(a)}~Median of per-pair ratios, LLM translation over algorithmic, with 95\%
bootstrap intervals. \textbf{(b)}~Mean number of verus runs per proof,
grouped by result.}
\label{fig:llm-alg-main}
\vspace{-0.5em}
\end{figure}
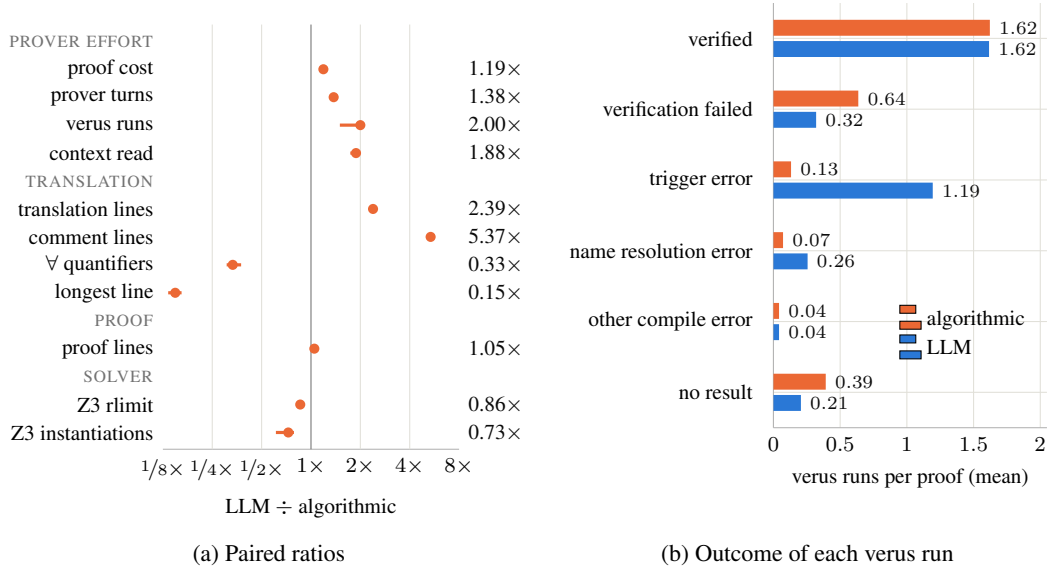

\begin{figure}[H]
  \centering
  %% GENERATED by fakellm-analysis-2026-09-09/gen_tex.py. Fragment: \input from the main document; data paths are relative to the project root.
% Fig 1 -- proof cost per problem, paired (x = algorithmic translation, y = LLM translation)
\begin{tikzpicture}
\begin{loglogaxis}[paired,
  xlabel={algorithmic translation, USD per proof}, ylabel={LLM translation, USD per proof},
  xmin=0.5, xmax=15, ymin=0.5, ymax=15,
  xtick={0.5,1,2,5,10}, ytick={0.5,1,2,5,10},
  xticklabels={\$0.5,\$1,\$2,\$5,\$10}, yticklabels={\$0.5,\$1,\$2,\$5,\$10}]
\addplot[gray!60] coordinates {(0.5,0.5) (15,15)};
\addplot[llmmark] table[x=alg_cost, y=llm_cost, col sep=comma] {balazs/data/fig1_cost.csv};
\node[anchor=north west, font=\small] at (rel axis cs:0.03,0.97) {LLM costlier in 71\% of pairs};
\node[anchor=south east, font=\small] at (rel axis cs:0.97,0.03) {Spearman $\rho = 0.71$};
\end{loglogaxis}
\end{tikzpicture}
  \caption{Proof cost per problem, paired. Each point is one (specification,
  property) pair.}
  \label{fig:llm-alg-cost}
\end{figure}
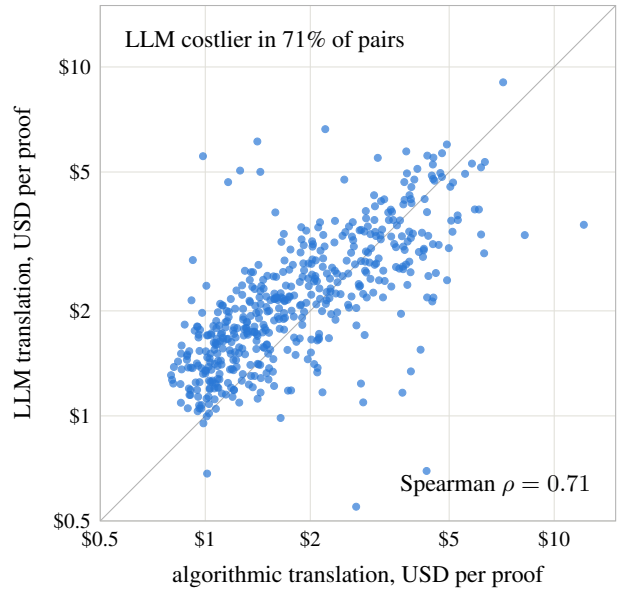

\begin{figure}[H]
  \centering
  %% GENERATED by fakellm-analysis-2026-09-09/gen_tex.py. Fragment: \input from the main document; data paths are relative to the project root.
% Fig 4 -- Z3 resource use per proof function: pooled ECDF of per-function rlimit (400-quantile downsample)
\begin{tikzpicture}
\begin{semilogxaxis}[base, width=9cm, height=6.5cm,
  xlabel={Z3 rlimit per proof function}, ylabel={fraction of proof functions},
  ymin=0, ymax=1, ytick={0,0.5,1}, yticklabels={0\%,50\%,100\%},
  xmin=1e3, xmax=1e8, legend pos=south east]
\addplot[alg, thick, const plot] table[x=rlimit, y=F, col sep=comma] {balazs/data/fig4_ecdf_rlimit_alg.csv};
\addlegendentry{algorithmic}
\addplot[llm, thick, const plot] table[x=rlimit, y=F, col sep=comma] {balazs/data/fig4_ecdf_rlimit_llm.csv};
\addlegendentry{LLM}
\end{semilogxaxis}
\end{tikzpicture}
  \caption{Z3 resource use per proof function.}
  \label{fig:llm-alg-rlimit}
\end{figure}
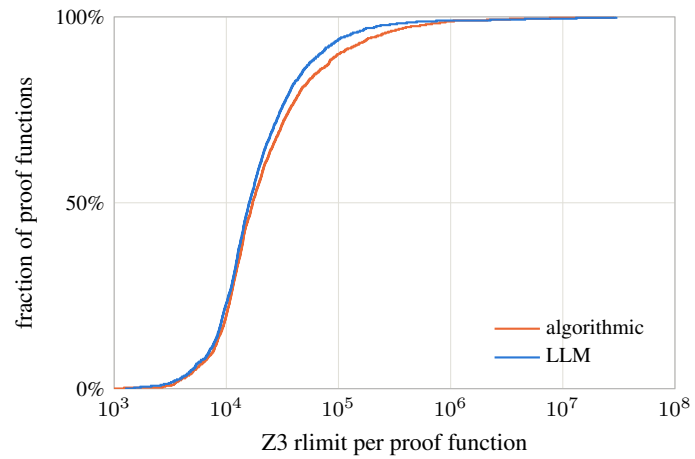

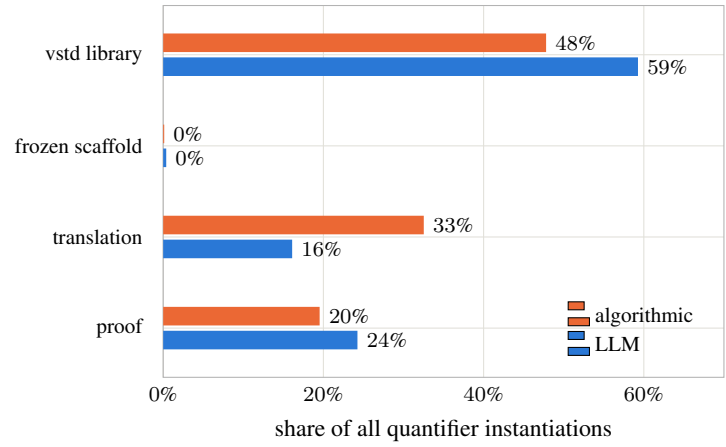
\begin{figure}[H]
  \centering
  %% GENERATED by fakellm-analysis-2026-09-09/gen_tex.py. Fragment: \input from the main document; data paths are relative to the project root.
% Fig 5 -- where Z3 instantiates quantifiers: share of all instantiations, summed over proofs
\begin{tikzpicture}
\begin{axis}[base, width=9cm, height=6.5cm, xbar, bar width=7pt,
  symbolic y coords={proof,translation,scaffold,vstd}, ytick={proof,translation,scaffold,vstd},
  yticklabels={proof,translation,frozen scaffold,vstd library},
  xmin=0, xmax=70, xtick={0,20,40,60}, xticklabel={\pgfmathprintnumber{\tick}\%},
  xlabel={share of all quantifier instantiations},
  nodes near coords={\pgfmathprintnumber[fixed,precision=0]{\pgfplotspointmeta}\%}, nodes near coords style={font=\footnotesize},
  legend pos=south east, reverse legend, enlarge y limits=0.18]
\addplot[fill=llm, draw=none] table[x expr=\thisrow{llm_share}*100, y=class, meta expr=\thisrow{llm_share}*100, col sep=comma] {balazs/data/fig5_inst_shares.csv};
\addlegendentry{LLM}
\addplot[fill=alg, draw=none] table[x expr=\thisrow{alg_share}*100, y=class, meta expr=\thisrow{alg_share}*100, col sep=comma] {balazs/data/fig5_inst_shares.csv};
\addlegendentry{algorithmic}
\end{axis}
\end{tikzpicture}
  \caption{Share of all Z3 instantiations by the
  location of the quantifier.}
  \label{fig:llm-alg-instantiations}
\end{figure}

\begin{figure}[H]
  \centering
  %% GENERATED by fakellm-analysis-2026-09-09/gen_tex.py. Fragment: \input from the main document; data paths are relative to the project root.
% Fig 7 -- quantifiers Verus had to pick triggers for, per bundle (minus the 6 in the frozen scaffold)
\begin{tikzpicture}
\begin{axis}[base, width=9cm, height=6.5cm, ybar, bar width=8pt,
  symbolic x coords={b0,b1,b2,b3,b4,b5}, xtick=data,
  xticklabels={0,1--2,3--5,6--10,11--20,$>$20},
  ymin=0, ymax=90, ytick={0,25,50,75}, yticklabel={\pgfmathprintnumber{\tick}\%},
  xlabel={auto-triggered quantifiers per bundle}, ylabel={share of bundles},
  legend pos=north east, enlarge x limits=0.12]
\addplot[fill=alg, draw=none] table[x=bin, y expr=\thisrow{alg_share}*100, col sep=comma] {balazs/data/fig7_auto_triggers.csv};
\addlegendentry{algorithmic}
\addplot[fill=llm, draw=none] table[x=bin, y expr=\thisrow{llm_share}*100, col sep=comma] {balazs/data/fig7_auto_triggers.csv};
\addlegendentry{LLM}
\end{axis}
\end{tikzpicture}
  \caption{Quantifiers for which Verus auto-triggered quantifiers,
  excluding the six in the frozen scaffold.}
  \label{fig:llm-alg-triggers}
\end{figure}
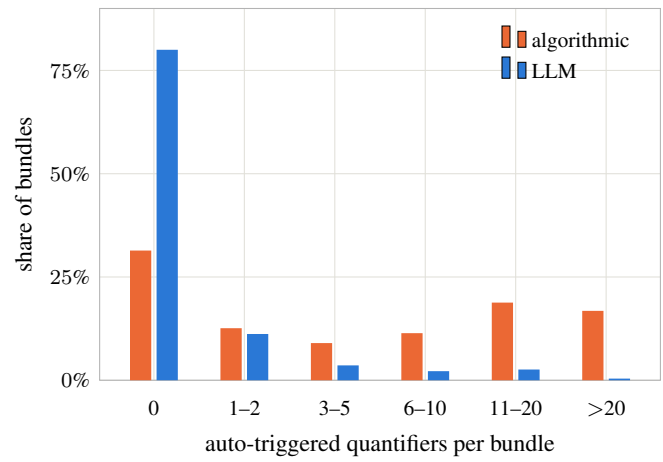

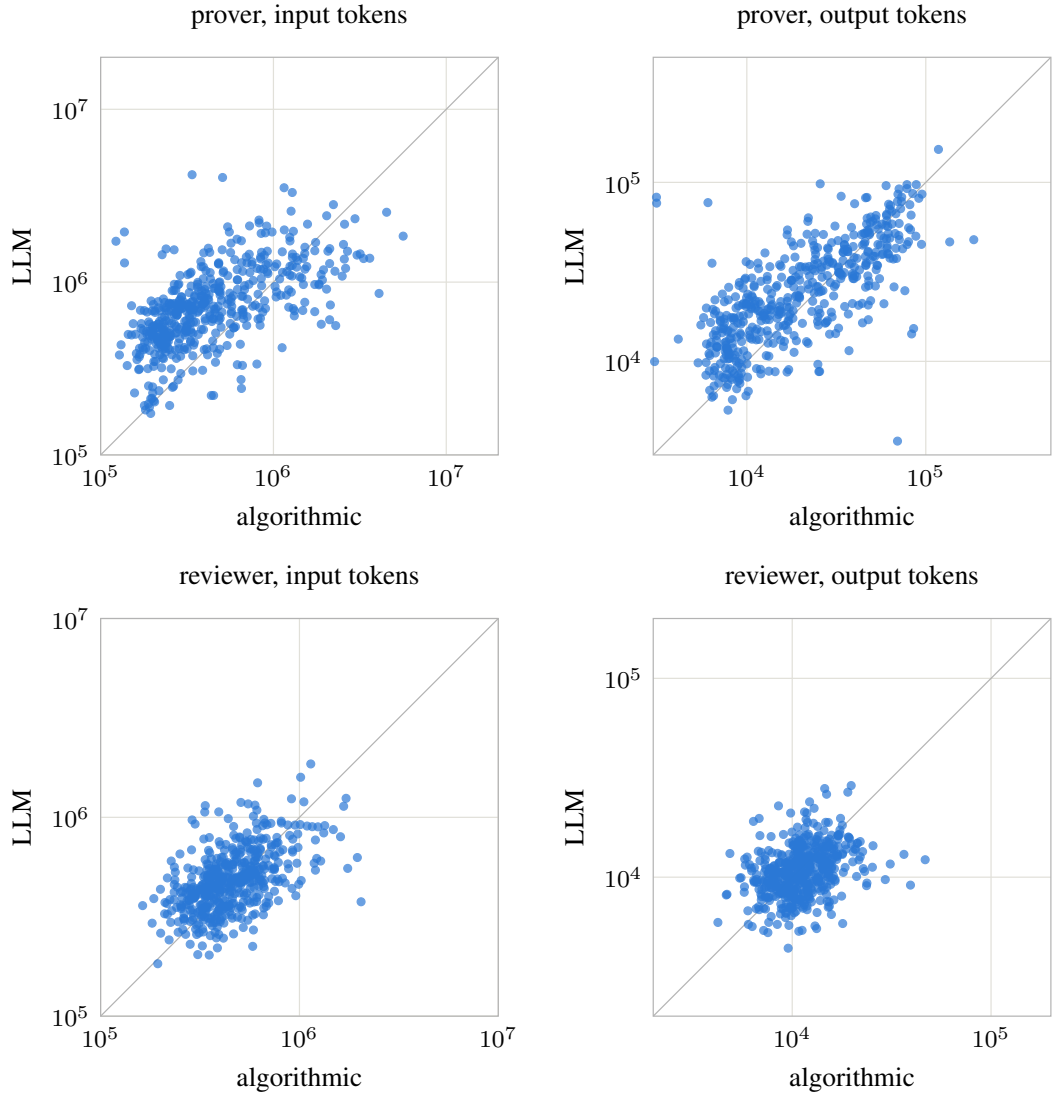
\begin{figure}[H]
  \centering
  \resizebox{\textwidth}{!}{%% GENERATED by fakellm-analysis-2026-09-09/gen_tex.py. Fragment: \input from the main document; data paths are relative to the project root.
% Fig 9 -- tokens per proof, paired (x = algorithmic, y = LLM; diagonal = equal). Input = cache read + cache write.
\begin{tikzpicture}
\begin{groupplot}[group style={group size=2 by 2, horizontal sep=1.8cm, vertical sep=1.9cm},
  base, width=6.2cm, height=6.2cm, xmode=log, ymode=log,
  xlabel={algorithmic}, ylabel={LLM}]
\nextgroupplot[title={prover, input tokens}, xmin=100000, xmax=2e+07, ymin=100000, ymax=2e+07]
\addplot[gray!60] coordinates {(100000,100000) (2e+07,2e+07)};
\addplot[llmmark] table[x=alg_prover_in, y=llm_prover_in, col sep=comma] {balazs/data/fig9_tokens.csv};
\nextgroupplot[title={prover, output tokens}, xmin=3000, xmax=500000, ymin=3000, ymax=500000]
\addplot[gray!60] coordinates {(3000,3000) (500000,500000)};
\addplot[llmmark] table[x=alg_prover_out, y=llm_prover_out, col sep=comma] {balazs/data/fig9_tokens.csv};
\nextgroupplot[title={reviewer, input tokens}, xmin=100000, xmax=1e+07, ymin=100000, ymax=1e+07]
\addplot[gray!60] coordinates {(100000,100000) (1e+07,1e+07)};
\addplot[llmmark] table[x=alg_reviewer_in, y=llm_reviewer_in, col sep=comma] {balazs/data/fig9_tokens.csv};
\nextgroupplot[title={reviewer, output tokens}, xmin=2000, xmax=200000, ymin=2000, ymax=200000]
\addplot[gray!60] coordinates {(2000,2000) (200000,200000)};
\addplot[llmmark] table[x=alg_reviewer_out, y=llm_reviewer_out, col sep=comma] {balazs/data/fig9_tokens.csv};
\end{groupplot}
\end{tikzpicture}}
  \caption{Tokens per proof, paired. Input counts
  cache reads and cache writes.}
  \label{fig:llm-alg-tokens}
\end{figure}

% ---------------------------------------------------------------------------

%%%%%%%%%%%%%%%%%%%%%%%%%%%%%%%%%%%%%%%%%%%%%%%%%%%%%%%%%%%%

\newpage
\section*{NeurIPS Paper Checklist}

\begin{enumerate}

\item {\bf Claims}
    \item[] Question: Do the main claims made in the abstract and introduction accurately reflect the paper's contributions and scope?
    \item[] Answer: \answerYes{} % Replace by \answerYes{}, \answerNo{}, or \answerNA{}.
    \item[] Justification: The contributions are listed explicitly at the end of \Cref{sec:intro}, each with a pointer to the section where it is developed.

\item {\bf Limitations}
    \item[] Question: Does the paper discuss the limitations of the work performed by the authors?
    \item[] Answer: \answerYes{} % Replace by \answerYes{}, \answerNo{}, or \answerNA{}.
    \item[] Justification: Limitations described \Cref{sec:discussion}.

\item {\bf Theory assumptions and proofs}
    \item[] Question: For each theoretical result, does the paper provide the full set of assumptions and a complete (and correct) proof?
    \item[] Answer: \answerNA{} % Replace by \answerYes{}, \answerNo{}, or \answerNA{}.
    \item[] Justification: The paper contains no theoretical results.

    \item {\bf Experimental result reproducibility}
    \item[] Question: Does the paper fully disclose all the information needed to reproduce the main experimental results of the paper to the extent that it affects the main claims and/or conclusions of the paper (regardless of whether the code and data are provided or not)?
    \item[] Answer: \answerYes{} % Replace by \answerYes{}, \answerNo{}, or \answerNA{}.
    \item[] Justification: The details of constructing our datasets can be found in \Cref{sec:phase1}, \Cref{sec:pahse2}, \Cref{sec:dataset}, \Cref{app:dataset_creation}, \Cref{app:reviewer}, \Cref{app:anticheat}. Information to reproduce experimental results can be found in \Cref{sec:setup} and \Cref{app:setup}.

\item {\bf Open access to data and code}
    \item[] Question: Does the paper provide open access to the data and code, with sufficient instructions to faithfully reproduce the main experimental results, as described in supplemental material?
    \item[] Answer: \answerNo{} % Replace by \answerYes{}, \answerNo{}, or \answerNA{}.
    \item[] Justification: This is a workshop paper on work in progress, we decided not to release the pipeline or the datasets before their final form.

\item {\bf Experimental setting/details}
    \item[] Question: Does the paper specify all the training and test details (e.g., data splits, hyperparameters, how they were chosen, type of optimizer) necessary to understand the results?
    \item[] Answer: \answerYes{} % Replace by \answerYes{}, \answerNo{}, or \answerNA{}.
    \item[] Justification: Experimntal settings can be found in great detail in \Cref{sec:setup} and \Cref{app:setup}.

\item {\bf Experiment statistical significance}
    \item[] Question: Does the paper report error bars suitably and correctly defined or other appropriate information about the statistical significance of the experiments?
    \item[] Answer: \answerNo{} % Replace by \answerYes{}, \answerNo{}, or \answerNA{}.
    \item[] Justification: We report point estimates without error bars. Each cell is k=3 samples per task on 40 tasks at temperature 1.0 with no seed fixed, and runs were not repeated, so sampling variability is not quantified. We therefore draw conclusions only from patterns that hold across models and across both encodings, and note this in \Cref{sec:discussion}.

\item {\bf Experiments compute resources}
    \item[] Question: For each experiment, does the paper provide sufficient information on the computer resources (type of compute workers, memory, time of execution) needed to reproduce the experiments?
    \item[] Answer: \answerYes{} % Replace by \answerYes{}, \answerNo{}, or \answerNA{}.
    \item[] Justification: Can be found in \Cref{app:compute}.
    
\item {\bf Code of ethics}
    \item[] Question: Does the research conducted in the paper conform, in every respect, with the NeurIPS Code of Ethics \url{https://neurips.cc/public/EthicsGuidelines}?
    \item[] Answer: \answerYes{} % Replace by \answerYes{}, \answerNo{}, or \answerNA{}.
    \item[] Justification: This paper presents work that aims to advance the field of Machine Learning. There are many potential societal consequences of our work, none of which we feel must be specifically highlighted here.

\item {\bf Broader impacts}
    \item[] Question: Does the paper discuss both potential positive societal impacts and negative societal impacts of the work performed?
    \item[] Answer: \answerNA{} % Replace by \answerYes{}, \answerNo{}, or \answerNA{}.
    \item[] Justification: This paper presents work that aims to advance the field of Machine Learning. There are many potential societal consequences of our work, none of which we feel must be specifically highlighted here.
    
\item {\bf Safeguards}
    \item[] Question: Does the paper describe safeguards that have been put in place for responsible release of data or models that have a high risk for misuse (e.g., pre-trained language models, image generators, or scraped datasets)?
    \item[] Answer: \answerNA{} % Replace by \answerYes{}, \answerNo{}, or \answerNA{}.
    \item[] Justification: We release no pre-trained models or generative systems. The dataset consists of TLA+ specifications from public repositories and their Verus encodings, i.e. formal specifications and machine-checked proofs, which carry no foreseeable misuse risk.

\item {\bf Licenses for existing assets}
    \item[] Question: Are the creators or original owners of assets (e.g., code, data, models), used in the paper, properly credited and are the license and terms of use explicitly mentioned and properly respected?
    \item[] Answer: \answerYes{} % Replace by \answerYes{}, \answerNo{}, or \answerNA{}.
    \item[] Justification: The sources and licensing are discussed in \cref{sec:dataset,app:setup}.

\item {\bf New assets}
    \item[] Question: Are new assets introduced in the paper well documented and is the documentation provided alongside the assets?
    \item[] Answer: \answerNA{} % Replace by \answerYes{}, \answerNo{}, or \answerNA{}.
    \item[] Justification: No assets are released at submission time. The dataset's
construction is documented in the paper (\Cref{sec:dataset}, \Cref{app:dataset-stats}).

\item {\bf Crowdsourcing and research with human subjects}
    \item[] Question: For crowdsourcing experiments and research with human subjects, does the paper include the full text of instructions given to participants and screenshots, if applicable, as well as details about compensation (if any)? 
    \item[] Answer: \answerNA{} % Replace by \answerYes{}, \answerNo{}, or \answerNA{}.
    \item[] Justification: No human subjects.

\item {\bf Institutional review board (IRB) approvals or equivalent for research with human subjects}
    \item[] Question: Does the paper describe potential risks incurred by study participants, whether such risks were disclosed to the subjects, and whether Institutional Review Board (IRB) approvals (or an equivalent approval/review based on the requirements of your country or institution) were obtained?
    \item[] Answer: \answerNA{} % Replace by \answerYes{}, \answerNo{}, or \answerNA{}.
    \item[] Justification: No human subjects.

\item {\bf Declaration of LLM usage}
    \item[] Question: Does the paper describe the usage of LLMs if it is an important, original, or non-standard component of the core methods in this research? Note that if the LLM is used only for writing, editing, or formatting purposes and does \emph{not} impact the core methodology, scientific rigor, or originality of the research, declaration is not required.
    %this research? 
    \item[] Answer: \answerYes{} % Replace by \answerYes{}, \answerNo{}, or \answerNA{}.
    \item[] Justification: Yes. LLMs are central to the method and their use is described throughout. \Cref{sec:phase1} details the agentic translation loop, in which a translator agent encodes TLA+ specifications into Verus and a reviewer agent checks faithfulness; \Cref{sec:pahse2} describes the prover--reviewer loop used for proof search; and \Cref{sec:modeleval} and \Cref{app:setup} give the eleven models evaluated on the resulting tasks, with the endpoints, decoding configuration and verification budgets used. Validity does not rest on model judgement: every accepted translation and proof is checked mechanically by Verus, with reviewer agents and anti-cheat checks (\Cref{app:anticheat}) applied on top.

\end{enumerate}

\end{document}